\documentclass[sensors,article,submit,moreauthors,pdftex,10pt,a4paper]{mdpi} 
\firstpage{1} 
\articlenumber{x}
\doinum{10.3390/------}
\pubvolume{xx}
\pubyear{2017}
\copyrightyear{2017}
\preto{\abstractkeywords}{\nolinenumbers}

\pdfoutput=1

\usepackage[section]{algorithm}
\usepackage{algpseudocode}
\usepackage{setspace}

\Title{Simultaneous Deployment and Tracking Multi-Robot Strategies with Connectivity Maintenance}

\Author{Javier Tardós $^{1}$, Rosario Aragüés $^{1}$*, Carlos Sagüés $^{1}$ and Carlos Rubio $^{1}$} 

\AuthorNames{Javier Tardós, Rosario Aragüés, Carlos Sagüés and Carlos Rubio}

\address[1]{%
	$^{1}$ \quad Universidad de Zaragoza, Instituto de Investigación en Ingeniería de Aragón; jtardos@outlook.com (J.T.); raragues@unizar.es (R.A.); csagues@unizar.es (C.S.);  carlosrr@unizar.es (C.R.)}

\corres{Correspondence: raragues@unizar.es; Tel.: +34 976 76 23 37}

\abstract{Multi-robot teams composed by ground and aerial vehicles have gained attention during the last years. We present a scenario where both types of robots must monitor the same area from different view points. In this paper we propose two Lloyd-based tracking strategies to allow the ground robots --agents-- follow the aerial ones --targets--, keeping the connectivity between the agents. The first strategy establishes density functions on the environment so that the targets acquire more importance than other zones, while the second one iteratively modifies the virtual limits of the working area depending on the positions of the targets. We consider the connectivity maintenance due to the fact that coverage tasks tend to spread the agents as much as possible, which is addressed by restricting their motions so that they keep the links of a Minimum Spanning Tree of the communication graph. We provide a thorough parametric study of the performance of the proposed strategies under several simulated scenarios. In addition, the methods are implemented and tested using realistic robotic simulation environments and real experiments.}


\begin{document}

\section{Introduction}
The coordination of autonomous robot teams has been a rising topic of interest during the last decades. Formation control, surveillance or coverage are some of the most faced topics. Heterogeneous robot teams are gaining attention for monitoring tasks in dynamic environments. For instance, in \cite{Stegagno-Tro2016} a method is presented for computing the localization of a group with both ground and aerial robots, using measurements that are not associated to identified robots. Considering the low battery life of the aerial robots, in \cite{Mathew-Tro2015} a group of them is assisted by a team of ground robots so that they can dock and recharge their batteries. Some authors focus on particular applications such as mobile sensory systems for monitoring environmental variables in greenhouses \cite{Roldan-Sens2016} instead of large sensor networks, or sediment sampling in estuarine mudflats \cite{Deusdado-Sens2016} with drilling ground vehicles and aerial imagery robots.

We face the problem of coordinating an heterogeneous team composed of both ground and aerial robots so that they cooperate to monitor an environment. This way, the different view points of the sensors can be used to build a richer representation of the 3D scene. In these scenarios, the design of the coordination strategy is challenging, since several restrictions should be simultaneously satisfied. Ground and aerial robots should observe the same area to build the representation of the scene. In order to cover the greater area, agents should be deployed far from each other. However, the communication should be kept so that they can exchange data and perform the computations. In this framework, we propose a strategy in which the aerial robots act as leaders of the formation. Ground robots --also referred to as agents-- execute different coverage-based algorithms to keep the aerial ones --from now on targets-- on their sensing range.

We consider two alternatives for the ground robots, both of them built on classical Centroidal Voronoi-based deployment ideas, which are also known as the Lloyd method, and has been often used for static deployment. The alternatives proposed in this paper face the adaptation of these ideas for tracking tasks. The first strategy consists of building density functions --also known as importance functions-- with focuses on the aerial robots. As they move, the density functions are adapted accordingly, making the ground robots track them. All of the ground agents need to know where the density functions are. This strategy has been previously used in the literature as it is immediately extracted from the Lloyd algorithm. In the second strategy, we build virtual boundaries around the regions in which the aerial and ground robots are operating. Ground robots coordinate to evenly deploy over the region enclosed by these time varying virtual boundaries. This strategy has never been used in the literature, as far as we know. It has the benefit that only the agents next to the boundaries need to be informed about them, and the others will react to their neighbors positions. For both strategies we ensure the ground robots remain connected by making them keep the links of a Minimum--distance Spanning Tree (MST) in the communication graph. This MST is recomputed at each iteration, so that agents have more freedom to move. Our method is centralized, although some of its steps admit a parallel execution.

We provide a thorough parametric study of the performance of the proposed strategies under several simulated scenarios built on MATLAB. We also evaluate the performance of the methods using realistic robotic simulation environments based on Gazebo and ROS (Robot Operating System), as well as using an experimental setup with six low-cost differential-drive  robots.

To sum up, the main contributions of this paper are: ($i$)~The proposal of methods for solving the problem of tracking a formation of targets or aerial robots with a team of agents or ground robots, covering the same areas as the targets, and keeping the network connected. ($ii$)~An algorithm based on virtual boundaries applicable to dynamic setups which makes agents to deploy in an equally spaced configuration. ($iii$)~We present extensive validations of our methods in different simulation scenarios and experiments under different setups.

This paper is organized as follows. Section \ref{related_work_section} gathers a review of related work in coverage, tracking and connectivity maintenance. In Section \ref{coverage_section} we present a base coverage algorithm with connectivity maintenance, using Lloyd method and MST, respectively. Then, in Section \ref{tracking_section} we propose two strategies to adapt the coverage algorithm for tracking tasks. These strategies are tested with MATLAB scenarios in Section \ref{simulations_section}. In Section \ref{realistic_sim_section} we carry out some realistic simulations with Gazebo and ROS. In Section \ref{experiment_section} we show the results of the real experiments. Finally, a set of conclusions and future work are given in Section \ref{conclusions_section}.

\section{Related Work}
\label{related_work_section}
As previously stated, we consider tasks of environmental monitoring where a team of ground and aerial robots establish coordination strategies to cover the same planar region with different view points. During their mission, ground agents must simultaneously satisfy several restrictions. They must follow the aerial robots and adapt dynamically to the positions covered by them. They also have to deploy in their coverage task, keeping the links of the communication network connected. Note that in this paper we will consider that coverage and monitoring tasks are similar, and will refer them indifferently. In this section, we give an overview of the strategies the most related to the situation considered in this paper, and discuss works on swarm aggregation, formation abstraction, formation control, containment control and leader-follower tracking. We discuss as well some applications of Voronoi and Lloyd methods.

Leader-follower methods \cite{Cao-Auto2009,Zhao-Auto2016} are consensus--based control laws that make a team of followers track the position of the leader. The aim is that robots remain and finish close to each other and to the leader. Swarm aggregation methods \cite{Dimarogonas-Tro2008,Leccese-ICRA2013} make robots move as a single entity through an environment, while avoiding obstacles. Although they do not collide, agents tend to be quite close to each other. In containment control tasks \cite{Cao-Auto2012,Kan-Auto2016,Shi-TAC2012}, there are multiple stationary or dynamic leaders, which are aware of the global objectives of the task. The remaining agents execute distributed local rules to follow the leaders, so that they converge to the convex hull spanned by the leaders. Another approach is formation abstraction \cite{Yoshida-tro2014}. The idea is to make agents remain inside the boundaries of a shape, e.g., polygonal, which can be modified or bent to adapt to the environment. The application is quite interesting, but ensuring collision avoidance or accurate tracking of the abstraction is usually hard to analyze. Here we consider a similar idea, but associated to the well known and widely used Lloyd method. In this context, it is easy to introduce other ideas such as connectivity maintenance and collision avoidance, as well as uniform distribution on the area. There are also other leader-follower behaviors of higher complexity, as \cite{Sabattini-tro2015}, where a team of leaders move so that they induce a set of periodic trajectories to be followed by the dependent robots. All these methods are not appropriate to deal with the problem we consider in this paper, since they may group too many agents, whereas we would like them to spread as much as possible to cover a larger area as they explore the environment. Moreover, some of the strategies described do not have control on the specific positions of the dependent robots.

Coverage and deployment algorithms are better suited for our purposes. The solutions considering Voronoi partitioning or the Lloyd method are the most common. They give rise to an easy to implement method, highly appealing. Voronoi partitioning consists of attaching to every robot the set of points of the working area that are closer to them, and is very popular on deployment and coverage tasks. In fact, many researches take advantage of Voronoi ideas. Some examples include \cite{Du-Auto2017}, where agents detect evaders within their Voronoi partition and pursue them. In \cite{Moon-ICUAS2017} the goal is to capture targets randomly distributed at an environment. Each agent detects targets inside its Voronoi cell and estimate their position with a Kalman Filter. The Lloyd method uses the Voronoi partitions to deploy the robots in the environment and they are iteratively moved to the centroid of their Voronoi cell, separating them from each other and spreading over the working area. This method is widely used in the literature. The situation where some areas have more importance, and robots have unlimited sensing and communication capabilities, can be solved \cite{Cortes-TRA04, Martinez-TAC07, Martinez-TAC07b} by iteratively moving the robots so that they reach centroidal Voronoi configurations. Several variations of this method have been presented to consider, e.g., robots using outdated information about the positions of the other team members \cite{Nowzari-AUTO2012}, or agents that can only sporadically send information to a central base station \cite{Patel-TCNS2016}. The works discussing limited sensing and communication capabilities are more realistic and thus have a higher interest. Circular sensing footprints are included in \cite{Cortes-ESAIM05}, while \cite{Pierson-IJRR2017} focuses on how to learn and adjust the regions associated to each robot depending on their sensing and actuating performances. Lloyd methods also allow the inclusion of obstacle-avoidance by the definition of safety regions that ensure there will be no collisions between the agents \cite{Kantaros-auto2016}. These previous Lloyd-based strategies focus on static environments, while we want to make the agents track the mobile aerial robots. Lloyd coverage methods have been also used to track moving targets. The most commonly used solution is to assign the density functions describing the importance of the environment to the targets, and make them change with time. It has been experimentally validated that the Lloyd method works properly in these cases \cite{Li-ICRA17,Lee-tro2015}. This is one of the strategies we also adopt in our work. However, in the presence of density functions, agents tend to accumulate nearby the targets. For this reason, we explore an alternative strategy, that makes the agents spread uniformly along a varying working region. 

A limitation of some multi-robot strategies is that often there are no guarantees that the network will remain connected as the agents operate. This is an important issue, since robots often must be able to exchange data and communicate with each other in order to coordinate and to successfully fulfill their tasks. In \cite{Gasparri-TRO2017} this term is included in the control law governing the global coordination objective. It keeps an accurate estimate of the time-varying Fiedler eigenvector. Depending on the high-level coordination strategy, the connectivity control term will affect its achievement. It is remarkable that they study both theoretically and experimentally these effects. In \cite{Boskos-JCO2017} there is a control law that enforces connectivity, and an additional bounded control term, that encapsulates the high-level control law for multi-robot cooperation. In addition, they design a control law that guarantees that the trajectories of the agents remain within a domain. This is done by means of a repulsion vector nearby the boundaries of the domain. A review of other connectivity control methods can be found in \cite{Zavlanos-ProcIEEE2011}. All these control methods often rely on keeping up to date estimates of data that encapsulates global properties.

An alternative to maintain connectivity is to make the agents keep a set of links in the network. The less restrictive method is to build a Tree in the graph. Specifically, the most used is the Minimum Spanning Tree (MST), where the links are selected depending on an assigned weight. Note also that as agents move, the weights of the tree branches change as well, and the last MST becomes obsolete. So, this MST should be periodically recomputed. Several works adopt this method for connectivity maintenance in different situations. A method for continuous-time systems that preserves a set of links is proposed in \cite{Li-CDC2009}. They mention that the best structure to be kept would be a Spanning Tree, but any strategy is discussed for adapting the Tree--links as robots move. In \cite{Schuresko-JCO2012} a high-level method is presented for managing Tree topologies which can be combined with motion tasks such as coverage, but the connections between the successive Trees and the Minimum Spanning Tree of the communication graph have not been established yet. In \cite{Aranda-16ACC} the authors present a formation control method for unicycle kinematics robots that keeps the connectivity. A triggered method for readjusting the Tree at some time instances during deployment tasks is developed in \cite{Aragues-14ECC}. In \cite{Soleymani-15ACC} the connectivity control takes place in a middle-ware layer to modify the goals of the robots when there is some risk of breaking some constraint. Also in \cite{Wagenpfeil-IFAC2009} the goal is to keep the links of a Tree, done in this case in a separate layer at constant time intervals.  In this paper we explore a different approach. We make the agents keep the links of the Minimum--distance Spanning Tree (MST) of the communication graph. So, we consider the re-computation of the MST at every time step, as it is done in \cite{Wagenpfeil-IFAC2009, Soleymani-15ACC}. However, in order to alleviate the communication load, a triggering strategy as the one proposed in \cite{Aranda-16ACC, Aragues-14ECC} could be used.

As a result, our methods are centralized. However, as acknowledged in \cite{Robin-AR2016}, some of the assumptions established for graph-based distributed control strategies are difficult to accommodate in realistic systems, and it may be reasonable to relax requirements on distributed methods. Some of the parts of our method are anyway highly parallelizable. For instance, the MST computation could be alleviated by using the distributed algorithm presented in \cite{Gallager-ACM_TPLS1983}. We will investigate in the future ways to transfer information on densities functions and on the virtual boundaries along the robot networks, to get an even more distributed method. Our aim is to get one step closer to a real implementation of the aerial-ground task, so we use a realistic simulation environment, as well as experiments with low-cost differential drive robots.

\section{Coverage for Static Deployment}
\label{coverage_section}

This section describes the basis of the coverage algorithm for a static environment. The ideas described in this section are a compound of the methods used in \cite{Cortes-TRA04, Martinez-TAC07, Martinez-TAC07b, Cortes-ESAIM05, Aragues-14ECC, Gallager-ACM_TPLS1983, Cao-tech02}. The coverage task is performed using the Lloyd method. It is appropriate to remark here that the algorithm described is implemented only in the ground robots.

\subsection{Static Deployment}
Considering a planar environment given by a convex polygon $Q \in \mathbb{R}^2$, we want to deploy $n$ agents with positions $p_i \in Q$, for $i=\{1,\dots,n\}$, in order to minimize the distances between them and any point $q \in Q$. This is known in the literature as a locational optimization problem, and here is solved using the Voronoi regions method. Partitioning $Q$ into $n$ regions, $\mathcal{W} = \left\{ W_1,...,W_n \right\}$, the coverage goal is accomplished by minimizing the cost function \cite{Cortes-TRA04}

\begin{equation}
\label{cost_function}
\mathcal{H} \left( P,\mathcal{W} \right) = \sum_{i=1}^n \int_{W_i} f \left( \left\| q-p_i \right\| \right) \phi \left( q \right) dq,
\end{equation}

\noindent where $f \left( \left\| q-p_i \right\| \right)$ describes the performance of the sensing devices of the agents, and $\phi(q)$ is a distribution density function that denotes the importance of each point. We assume $\mathcal{W}$ as the Voronoi partition, $\mathcal{V} = \left\{ V_1,...,V_n \right\}$, where 

\begin{equation*}
V_i = \left\{ q \in Q \ | \ \left\| q-p_i \right\| \le \left\| q-p_j \right\|, \ \forall j \ne i \right\},
\end{equation*}

\noindent i.e., the regions defined by attaching to each node, $p_i$, their closest points, $q$. So, the cost function (Eq.~\ref{cost_function}) to minimize is now

\begin{equation}
\label{cost_function_2}
\mathcal{H} \left( P,\mathcal{V} \right) = \sum_{i=1}^n \int_{V_i} f \left( \left\| q-p_i \right\| \right) \phi \left( q \right) dq.
\end{equation}

The sensors of the agents observe better nearby points and, considering this fact, their performance function is defined as $f \left( \left\| q-p_i \right\| \right) = \left\| q-p_i \right\|^2$. Recalling some basic quantities associated to a region $W$ and a density function $\phi$, the mass $M_W$, center of mass $C_W$ and polar moment of inertia $J_{W,p}$ are defined as

\begin{equation*}
M_W = \int_{W} \phi \left( q \right) dq, \qquad \qquad
C_W = \frac{1}{M_W}  \int_{W} q \ \phi \left( q \right) dq, \qquad \qquad
J_{W,p} = \int_{W} \left\| q-p \right\|^2 \phi \left( q \right) dq.
\end{equation*}

Considering the parallel axis theorem $J_{W,p} = J_{W,C_W} + M_W \left\| p-C_W \right\|^2$, the cost function (Eq.~\ref{cost_function_3}) and its gradient (Eq.~\ref{gradient}) can be calculated as

\begin{equation}
\label{cost_function_3}
\mathcal{H} \left( P \right) = \sum_{i=1}^n J_{V_i,C_{V_i}} + \sum_{i=1}^n M_{V_i} \left\| p_i-C_{V_i} \right\|^2
\end{equation}

\begin{equation}
\label{gradient}
\frac{\partial\mathcal{H}}{\partial p_i} \left( P \right) = 2 M_{V_i} \left( p_i-C_{V_i}\right),
\end{equation}

\noindent as done in \cite{Cortes-TRA04}. Thus, in order to deploy the agents so as to optimize the cost function, they iteratively compute their Voronoi regions and move to their center of mass (Eq.~\ref{center_mass}). This gradient descent method is commonly known as Lloyd method.

\begin{equation}
\label{center_mass}
C_{V_i} = \left( \int_{V_i} \phi \left( q \right) dq \right) ^{-1}  \int_{V_i} q \ \phi \left( q \right) dq
\end{equation}

\begin{figure}[!b]
	\centering
	\includegraphics[width=0.45\textwidth]{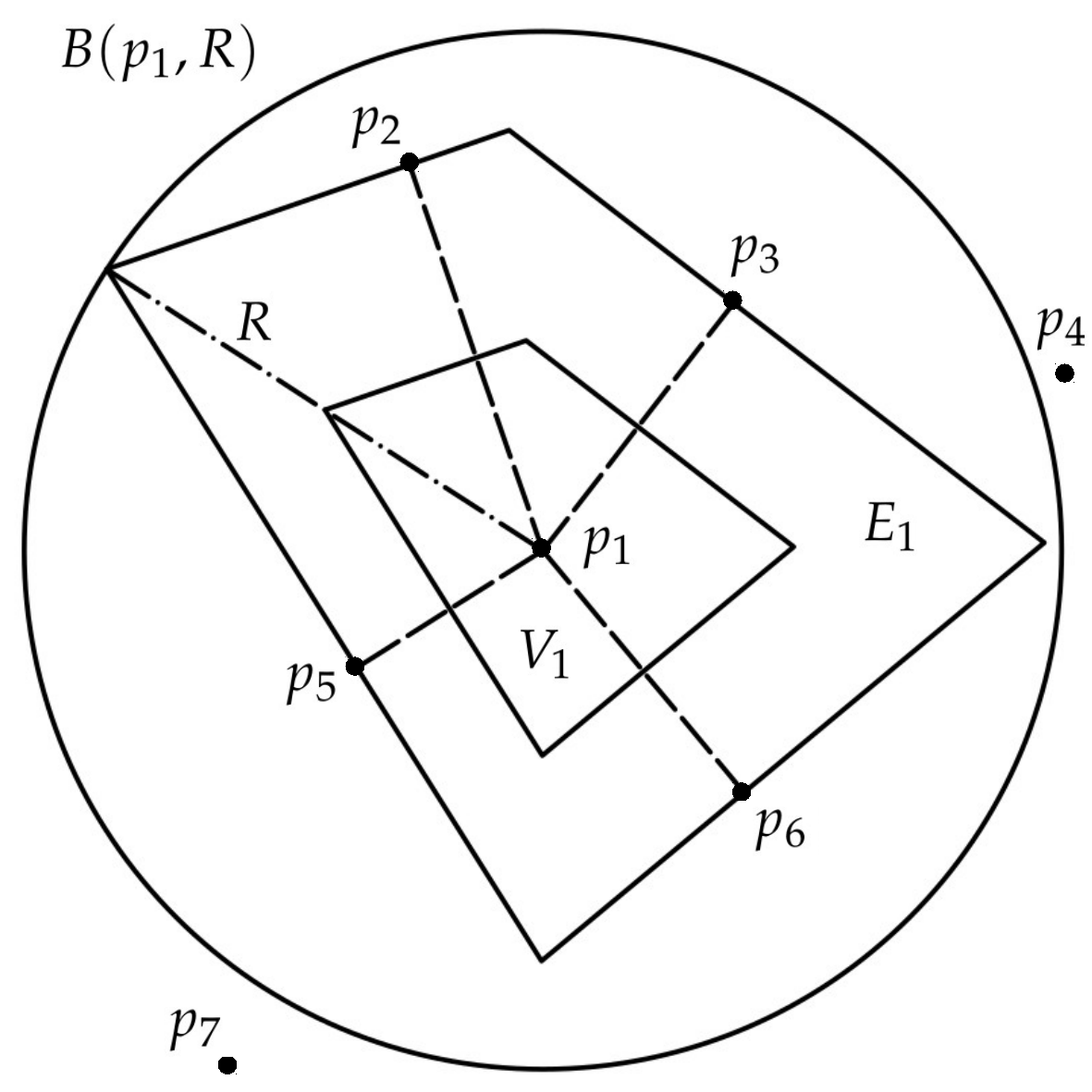}
	\caption{Criterion for selecting the Voronoi neighbors.}
	\label{cao_neighbors}
\end{figure}

Note that before the calculation of the Voronoi region of each agent, we need to know which agents are their neighbors. The process we follow to solve this problem is the simple criterion proposed in \cite{Cao-tech02}, explained next for the node $p_1$ (see Figure \ref{cao_neighbors}). Assuming that we know the position of all the nodes, $p_i$, the method starts computing the boundaries of the Voronoi region, $V_1$, by the nearest neighbor, then the second nearest, etc. After incorporating each one of these closest nodes, the algorithm checks whether there is any other node inside a circle $B(p_1,R)$, where $R$ is a changing radius equal to the maximum distance from $p_1$ to any point of the region $E_1$ computed until that iteration. This region is the one composed from drawing parallel lines to the Voronoi boundaries through the corresponding already found neighbors. If there are not any more agents inside this circle, all the neighbors have been found. In the example shown, $p_2$, $p_3$, $p_5$ and $p_6$ are Voronoi neighbors of $p_1$, while $p_4$ and $p_7$ are not.

\subsection{Limited Sensing}
The sensors carried by the agents have a limited range, so that they can not correctly observe objects outside a circle with center $p_i$ and radius, their sensing radius $s$. Thus, instead of the performance sensing model $f \left( \left\| q-p_i \right\| \right) = \left\| q-p_i \right\|^2$, we have a sensing area limited by $s$ \cite{Aragues-14ECC}.

\begin{align*}
f(\|q-p_i\|)&=\left\{ 
\begin{array}{ll} \|q-p_i\|^2, &\mathrm{if~} \|q-p_i\|< s,\\
s^2, &\mathrm{otherwise}. \end{array} \right. 
\end{align*}

\noindent Therefore, the coverage region of each agent is the intersection between the sensing area of each agent, $B(p_i,s)$, and its corresponding Voronoi partition, $V_i$. As a result, the cost function associated to this limitation is

\begin{equation}
\label{cost_function_s}
\mathcal{H}_s \left( P \right) = \sum_{i=1}^n \int_{V_i \cap B(p_i,s)} \left\| q-p \right\|^2 \phi \left( q \right) dq.
\end{equation}

\noindent And the gradient of $\mathcal{H}_s$ in this case is

\begin{equation}
\label{gradient_s}
\frac{\partial\mathcal{H}}{\partial p_i} \left( P \right) = 2 M_{V_i \cap B(p_i,s)} \left( p_i-C_{V_i \cap B(p_i,s)}\right),
\end{equation}

\noindent i.e., the gradient descent move to the centroid, but of the region defined as the intersection between the Voronoi region and the sensing area.

Referring to Proposition 2.7 in \cite{Cortes-ESAIM05}, a local minimum of this limited cost function (Eq.~\ref{cost_function_s}), $P^*=(p_1^*,\dots,p_n^*)$, is a local minimum of the cost function without limited range interactions (Eq.~\ref{cost_function_2}) if $Q \subset \cup_{i \in \{1,\dots,n\}}B(p_i,s)$. So, this approximation obtains the local optimum solution of the locational problem considering only the area covered by the sensing radius of the agents.

\subsection{Limited Communication}
For the communication purposes, we consider an \textit{r-limited Delaunay graph}, as defined in \cite{Cortes-ESAIM05}, with a communication radius $r$. In order to be able to run the methods proposed in this section in a distributed fashion, the \textit{r-limited Delaunay graph} requires $r\geq 2s$ communication radius, knowing the positions of the agents at a distance at least $2s$. If our communication radius is smaller, we need to consider an altered cost function in which we will reduce the sensing region. Suppose $s^*$ is our original sensing radius, and $s$ the reduced one, so that $2s\leq r$. Following the steps in Proposition 2.5 \cite{Cortes-ESAIM05}, we can relate their corresponding limited cost functions $\mathcal{H}_{s^*}$ and $\mathcal{H}_s$. Let $f(x)=x^2$ be our performance function, where $f(0)=0$. Since we have a limited sensing problem, the performance function $f_{s^*}$ is defined as $f_{s^*}(x)=f(x)$ for $x<s^*$, and $f_{s^*}(x)=f(s^*)$ for $x\geq s^*$. Reducing the sensing radius for our distributed purposes, for $s \in \left[0, s^*\right]$, we define the performance function $f_s$ given by $f_s(x)=f(x)$ for $x<s$, and $f_s(x)=f(s)$ for $x\geq s$. Considering that $s^* \geq s$, we also define a constant $\beta=\frac {f\left(s^*\right)} {f(s)} \geq 1$, and setting $b=f\left(s^*\right)$ then $f_{s^*}(x)\leq b, \ \forall \ x \in \left[0, s^*\right]$. 

By construction, $f_s(x)\leq f_{s^*}(x), \ \forall \ x \in \left[0, s^*\right]$, so we can conclude that $\mathcal{H}_s(P) \leq \mathcal{H}_{s^*}(P)$. Now, consider the function $\tilde{f}(x)=\beta f_s(x)$. Note that $\tilde{f}(x)=\beta f(x)\geq f_{s^*}(x)$ for $x<s$, and $\tilde{f}(x)=\beta f(s)=b=f_{s^*}(x)$ for $x\geq s$. Therefore, we also can conclude that $\beta \mathcal{H}_s(P) \geq \mathcal{H}_{s^*}(P)$ and then, for all $P\in \{\cup^n_{i=1}B(p_i,s^*)\}^n$,

\begin{equation}
\label{costfnc_limit}
\beta \mathcal{H}_s(P) \geq \mathcal{H}_{s^*}(P) \geq \mathcal{H}_s(P) > 0
\end{equation}

Thus, as we can see, reducing the sensing radius so that this part of the method can be run on 1-hop neighbors, gives rise to another cost function, with specific relations to the original one.

\subsection{Particular Closed-Form Expressions for Constant Density Functions}
When the density function, $\phi(q)$, is defined as a constant, the centers of mass of the coverage regions of each agent coincide with their geometric centroids. Assuming that the coverage regions, $W_i$, are convex polygons with $N_i$ vertexes labeled as $\left\{ \left( x_0,y_0 \right),..., \left( x_{N_i-1},y_{N_i-1} \right) \right\}$, the geometric centroid is calculated as \cite{Cortes-TRA04}

\begin{equation}
\label{centroid_noImp}
\begin{aligned}
C_{W_i,x} = \frac{1}{6M_{W_i}} \sum_{k=0}^{N_i-1} \left( x_k + x_{k+1} \right) \left( x_ky_{k+1} - x_{k+1}y_k \right)\\
C_{W_i,y} = \frac{1}{6M_{W_i}} \sum_{k=0}^{N_i-1} \left( y_k + y_{k+1} \right) \left( x_ky_{k+1} - x_{k+1}y_k \right),
\end{aligned}
\end{equation}

\noindent where

\begin{equation}
M_{W_i} = \frac{1}{2} \sum_{k=0}^{N_i-1} \left( x_ky_{k+1} - x_{k+1}y_k \right).
\end{equation}

\section{Simultaneous Deployment and Tracking with Connectivity Maintenance}
\label{tracking_section}

We introduce an explanation of the connectivity maintenance method and  we give details on the control law. After that, we explain our two alternative methods for simultaneous deployment and tracking.

\subsection{Connectivity Maintenance}
\label{section_connCtrl}
Regarding the connectivity maintenance problem, we use a Minimum Spanning Tree (MST)  \cite{Aranda-16ACC,Aragues-14ECC, Soleymani-15ACC,Wagenpfeil-IFAC2009}. This method joints the $n$ agents with $n-1$ links. The nodes are connected in a communication tree that minimizes the sum of the lengths of the links in order to allow the maximum motion freedom. The MST must be recalculated at each step of the simulation. Otherwise, the links would heavily restrict the motion as agents move along. With all these considerations, the process that the algorithm runs at each iteration to calculate the MST in a centralized version is shown in the Algorithm \ref{MST_algorithm}. A decentralized version can be found in \cite{Gallager-ACM_TPLS1983} and a triggered strategy is shown in \cite{Aragues-14ECC}.

As mentioned above, the MST is used to ensure the network remains connected as agents move. Observe Algorithm \ref{Lloyd_1step_connCtrl}, which includes the main statements in one iteration of the Lloyd algorithm with connectivity maintenance. At every iteration $k$, each agent $i$ computes the Centroid $C_{W_i}(k)$ of its Voronoi region, and computes its next position as
\begin{align}
p_i(k+1)&=p_i(k)+u_i(k), &u_i(k)=K(C_{W_i}(k) - p_i(k)).
\label{eq_controlLaw1}
\end{align}
If there are no restrictions on the agent motion, then $K=1$ and the next agent position $p_i(k+1)$ is its current centroid, $p_i(k+1)=C_{W_i}(k)$.

\begin{algorithm}[!t]
	\caption{MST calculation} 
	\label{MST_algorithm}
	\begin{doublespace}
		\begin{algorithmic}[1]
			\State Distances$(i,j)=\left\|p_i-p_j\right\|$
			\State Communicables$(i,j)=\text{Distances}(i,j)\leq r$
			\State MST\_Graph $=zeros(n\text{x}n)$
			\State MST\_Nodes $=1$
			\While {$length(\text{MST\_Nodes})\leq(n-1)$}
			\ForAll {$i\in$ MST\_Nodes}
			\State Candidates $= \ find \ (\text{Communicables}(\text{MST\_Nodes}(i),:))$
			\ForAll {$j\in$ Candidates}
			\If {$\exists$ MST\_Nodes $=$ Candidates$(j)$ }
			\State MST\_Graph$(\text{MST\_Nodes}(i),\text{Candidates}(j))=0$
			\State MST\_Graph$(\text{Candidates}(j),\text{MST\_Nodes}(i))=0$
			\Else 
			\State Curr\_Distance $=\text{Distances}(\text{MST\_Nodes}(i),\text{Candidates}(j))$
			\If {Curr\_Distance $<$ Best\_Distance }
			\State Best\_Distance $=$ Curr\_Distance
			\State Best\_Candidate $=$ Candidates$(j)$
			\State Best\_Source $=$ MST\_Nodes$(i)$
			\EndIf
			\EndIf
			\EndFor
			\EndFor
			\State MST\_Nodes $= \ append(\text{Best\_Candidate})$
			\State MST\_Graph$(\text{Best\_Source},\text{Best\_Candidate})=1$
			\State MST\_Graph$(\text{Best\_Candidate},\text{Best\_Source})=1$
			\EndWhile
		\end{algorithmic}
	\end{doublespace}
\end{algorithm}

\begin{figure}[H]
	\centering
	\includegraphics[width=0.2\textwidth]{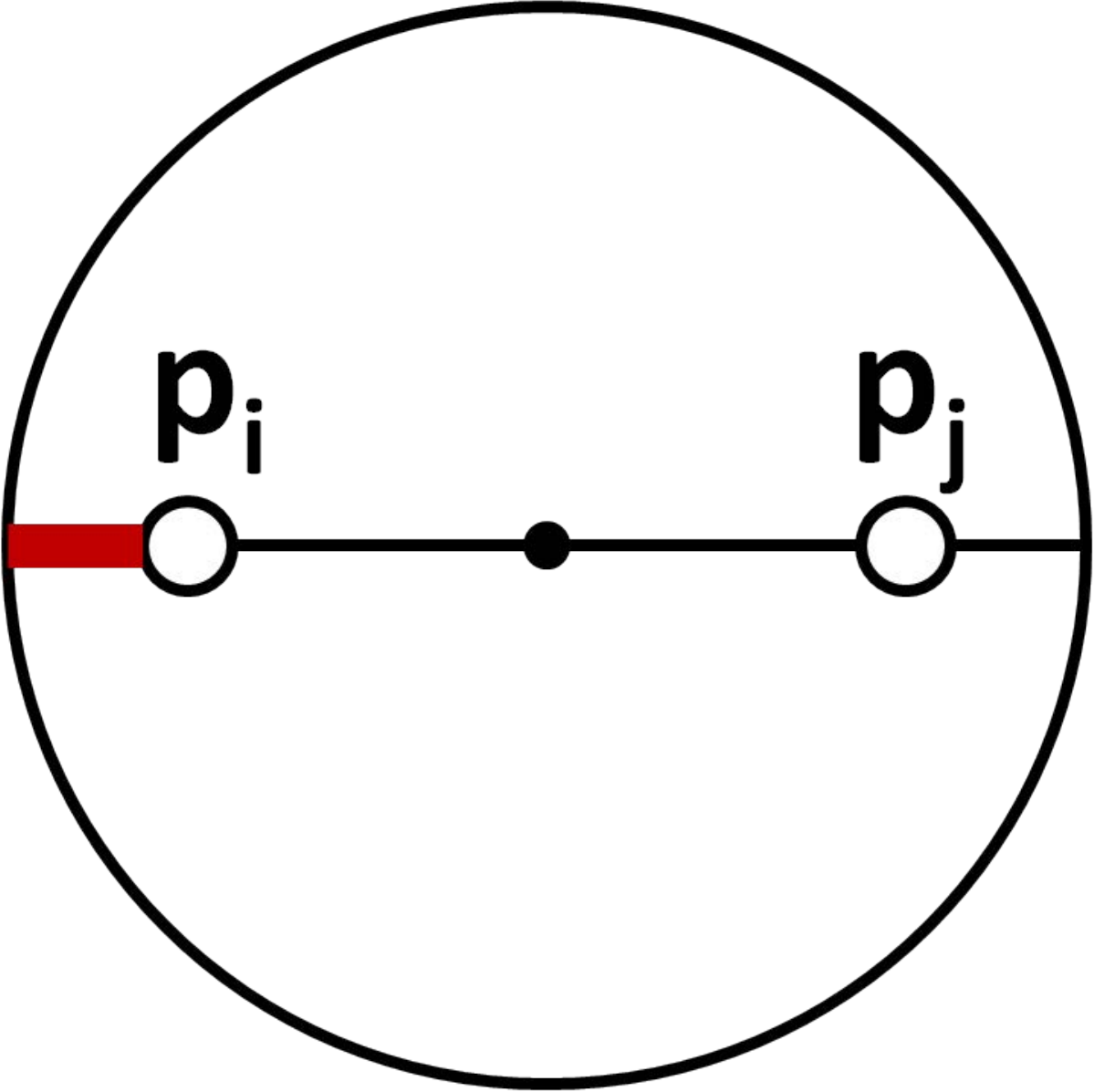}
	\caption{Motion restrictions for linked agents.}
	\label{motion_restr}
\end{figure}

In order to keep the connectivity of the whole system, each agent must be within the communication radius $r$ of its MST--neighbors. 
However, the motion of the ground robots towards the centroids of their coverage regions may break the communication links, if no additional restrictions are included. At every step, the algorithm has to limit the motion of each pair of nodes $(i,j)$ that are linked in the MST. When next positions $p_i(k+1), p_j(k+1)$ are restricted to a circle with center at  $(p_i(k)+p_j(k))/2$ and radius $r$, their distance remains within $r$~\cite{Ando-TRA1999}. Figure \ref{motion_restr} shows the maximum movement (red wide line) available in the worst case, where agents need to move in opposite directions.

Every $(i,j)$ link in the MST that must be kept for agent $i$, can introduce a restriction on its  next position $p_i(k+1)$. Agent $i$ adjust its control input gain $K$ in equation~\ref{eq_controlLaw1} so that $p_i(k+1)$ equals the closest point from $p_i(k)$ to $C_{W_i}(k)$ that  satisfies all the restrictions.

Note that agents may temporarily miss MST links as they navigate from $p_i(k)$ to $p_i(k+1)$. However, all these MST links are re-established as all robots get to their goal positions $p_i(k+1)$, which is when they really need the network to be connected since they perform operations that require the exchange of data.

\begin{algorithm}[H]
	\caption{Lloyd\_MSTConnCtrl} 
	\label{Lloyd_1step_connCtrl}
	\begin{doublespace}
		\begin{algorithmic}[1]
			\State Detect $(\text{Neighbors})$ \Comment {\cite{Cao-tech02}}
			\State Coverage\_regions $=\text{Voronoi\_regions}(\text{Neighbors}) \cap \text{Sensing\_regions}(\text{Neighbors},s) \cap Q$
			\State Compute centroid $C_{W_i} = \left( \int_{W_i} \phi \left( q \right) dq \right) ^{-1}  \int_{W_i} q \ \phi \left( q \right) dq$ \Comment {Eq. ~\eqref{center_mass}}
			\State Compute $(\text{MST\_Graph})$ \Comment {Algorithm \ref{MST_algorithm}}
			\State Goal\_Positions $=$ Limit$(C_{W_i},\text{MST\_Graph})$ \Comment {Section~\ref{section_connCtrl}}
			\State Navigate $(\text{Goal\_Positions})$
			\State Send $(\text{New\_Positions})$
		\end{algorithmic}
	\end{doublespace}
\end{algorithm}

Note that Algorithm~\ref{Lloyd_1step_connCtrl} generates goal positions for each agent $i=1,…,n$,  as if agents had integrator dynamics in a discrete-time setup. Our methods can anyway be used by  agents with other kinematic models. The positions $p_i(k+1)$ generated by our methods are high level control laws. Agents are equipped with local motion controllers, specific for their kinematic  models, to drive them nearby the high level goals $p_i(k+1)$. When they reach the goals with precision enough, agents inform each other and synchronize to start a new iteration of our algorithms. A detailed discussion of this control strategy can be found in Section~\ref{experiment_section}, for a team of six robot with differential drive kinematics.

\subsection{Varying Importance Functions}
We propose two alternatives to  deal with tracking tasks. The first alternative solves the tracking issue through the density function $\phi(q)$ introduced in the coverage algorithm. This approach has been previously used in the literature to set up a dynamic coverage from the Voronoi tessellation, as it is immediate from Lloyd static method. Each target is given an importance function defined as

\begin{equation}
\label{imp_function}
\phi \left( q \right) = e ^ {-\left\| q-o \right\|^2},
\end{equation}

\noindent where $o$ is the position of the target. These values weight the calculation of the centroids forcing the agents to reach the center of each function, where the targets place. Therefore, all the agents must be informed about the positions of all targets. Varying the position of these functions depending on the targets motion, we can achieve a tracking behavior. The ground agents will deploy covering the same area as the aerial robots.

\begin{figure}[H]
	\centering
	\begin{tabular}{cc}
		\includegraphics*[width=0.45\textwidth]{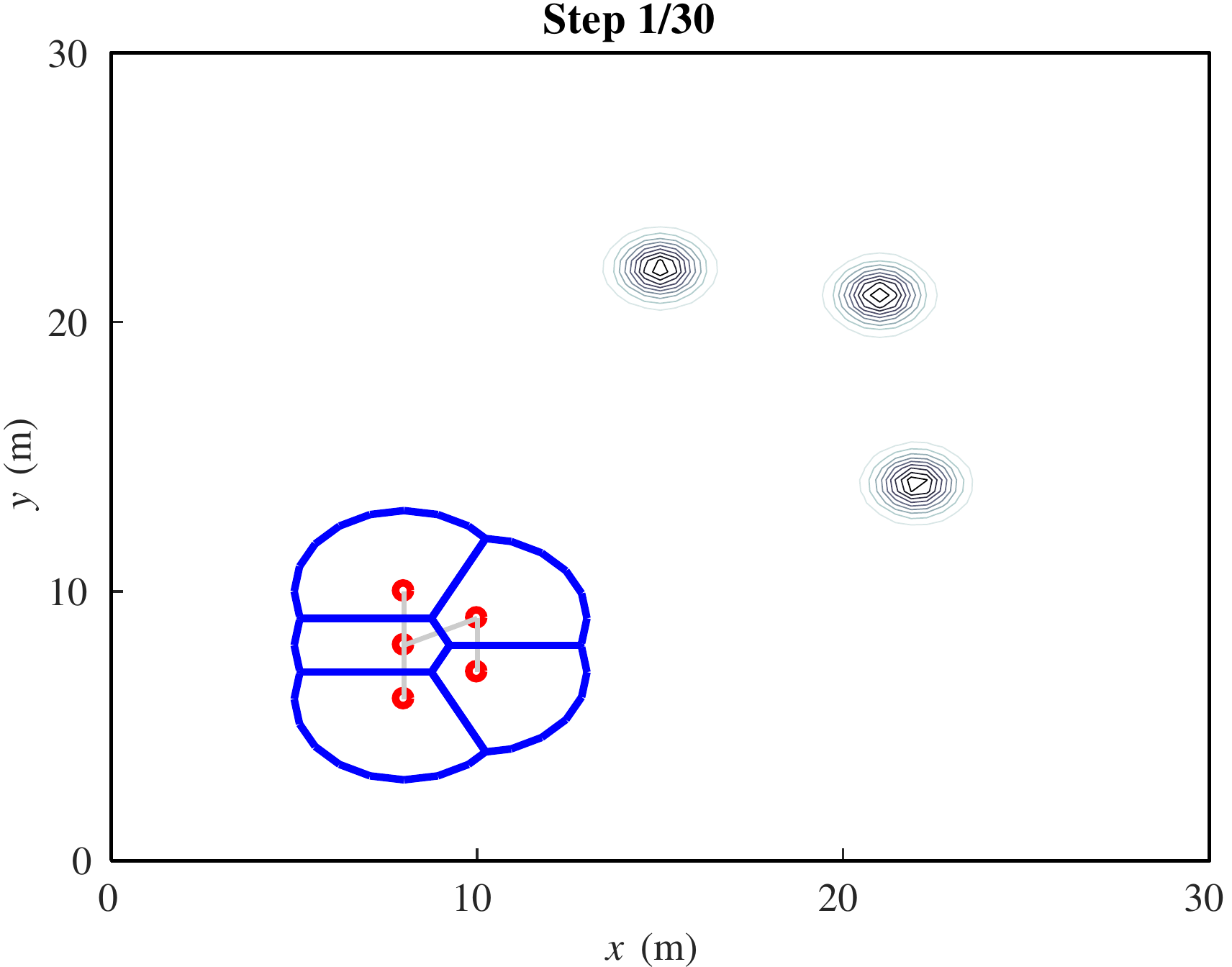}&
		\includegraphics*[width=0.45\textwidth]{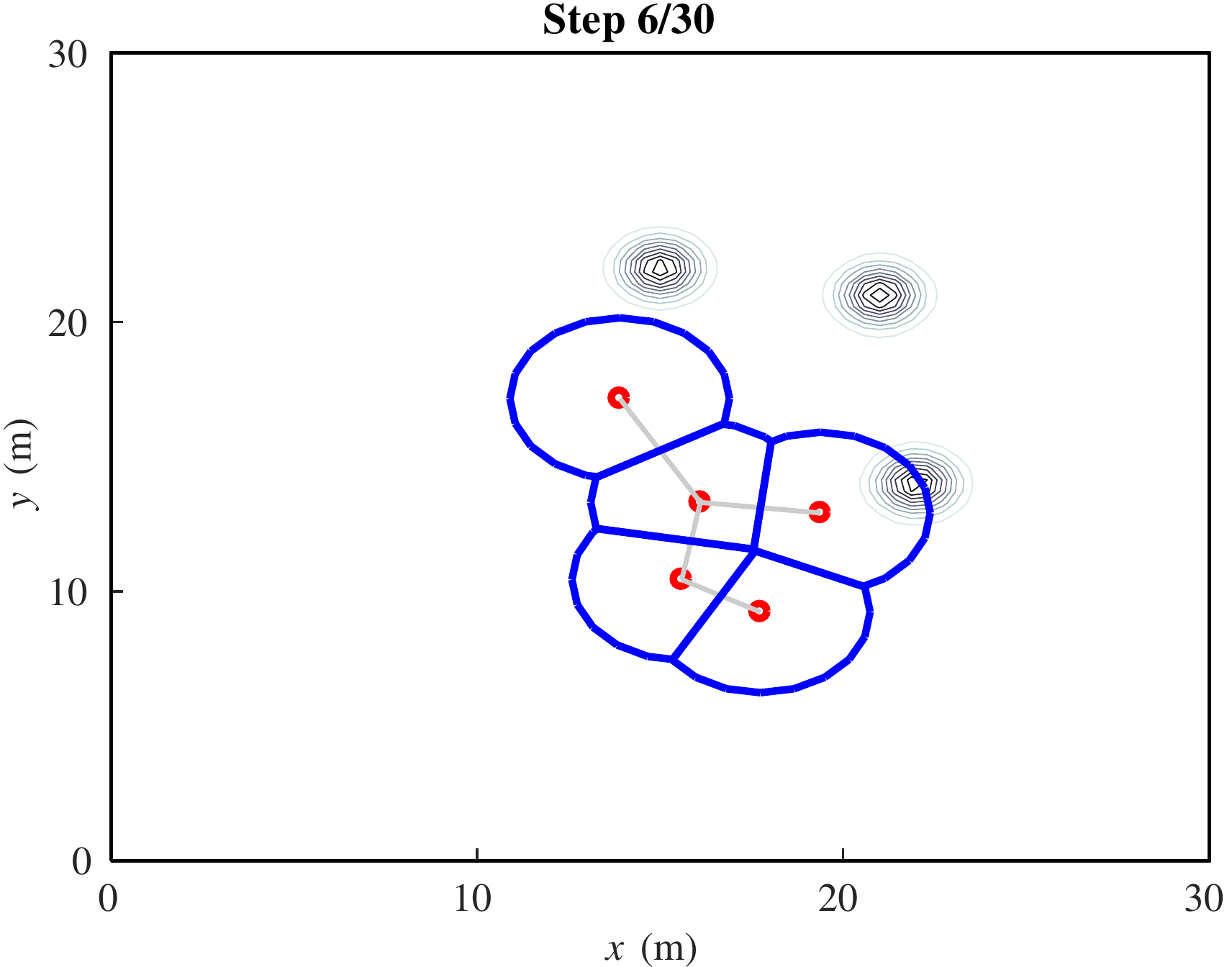}\\
		\includegraphics*[width=0.45\textwidth]{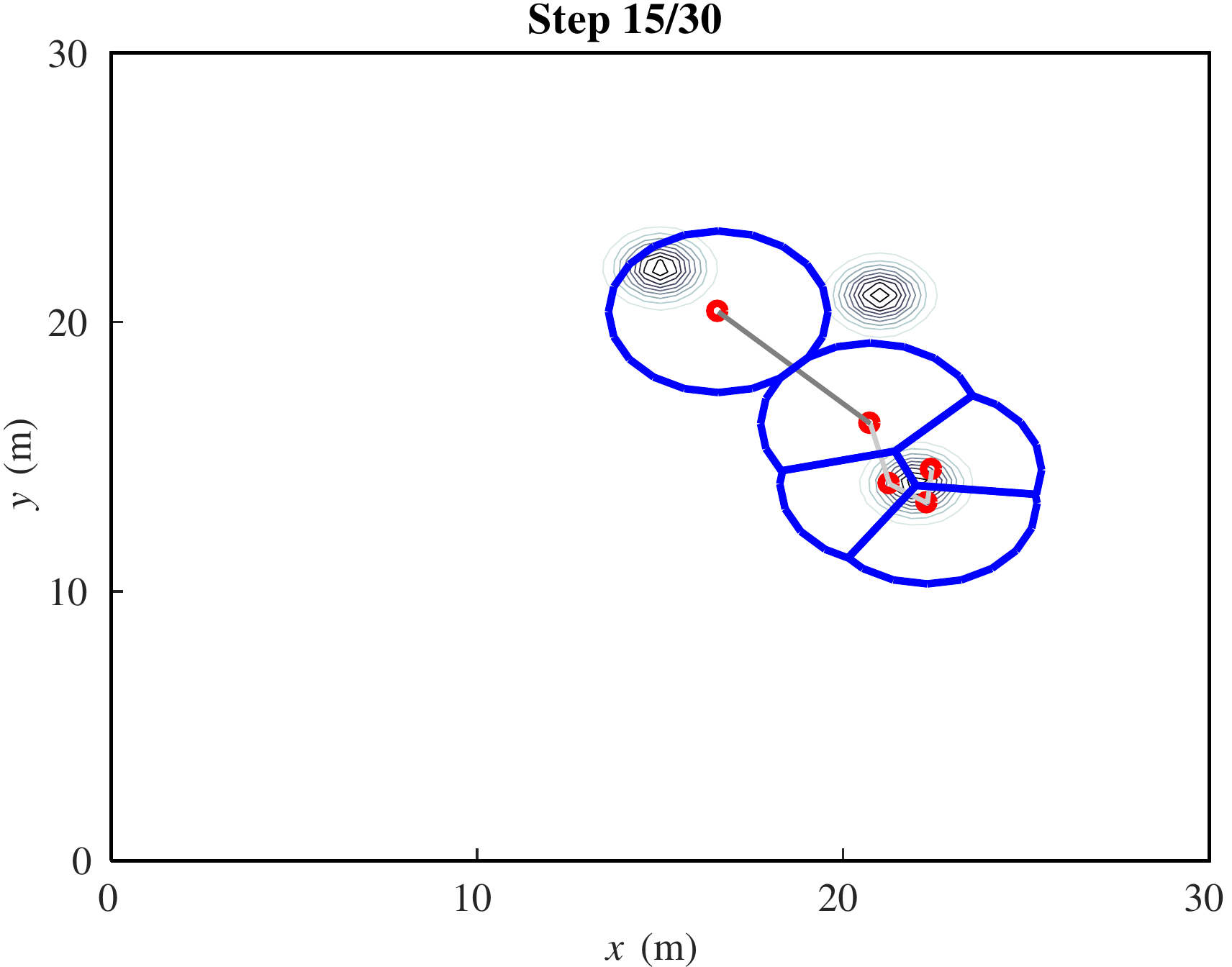}&
		\includegraphics*[width=0.45\textwidth]{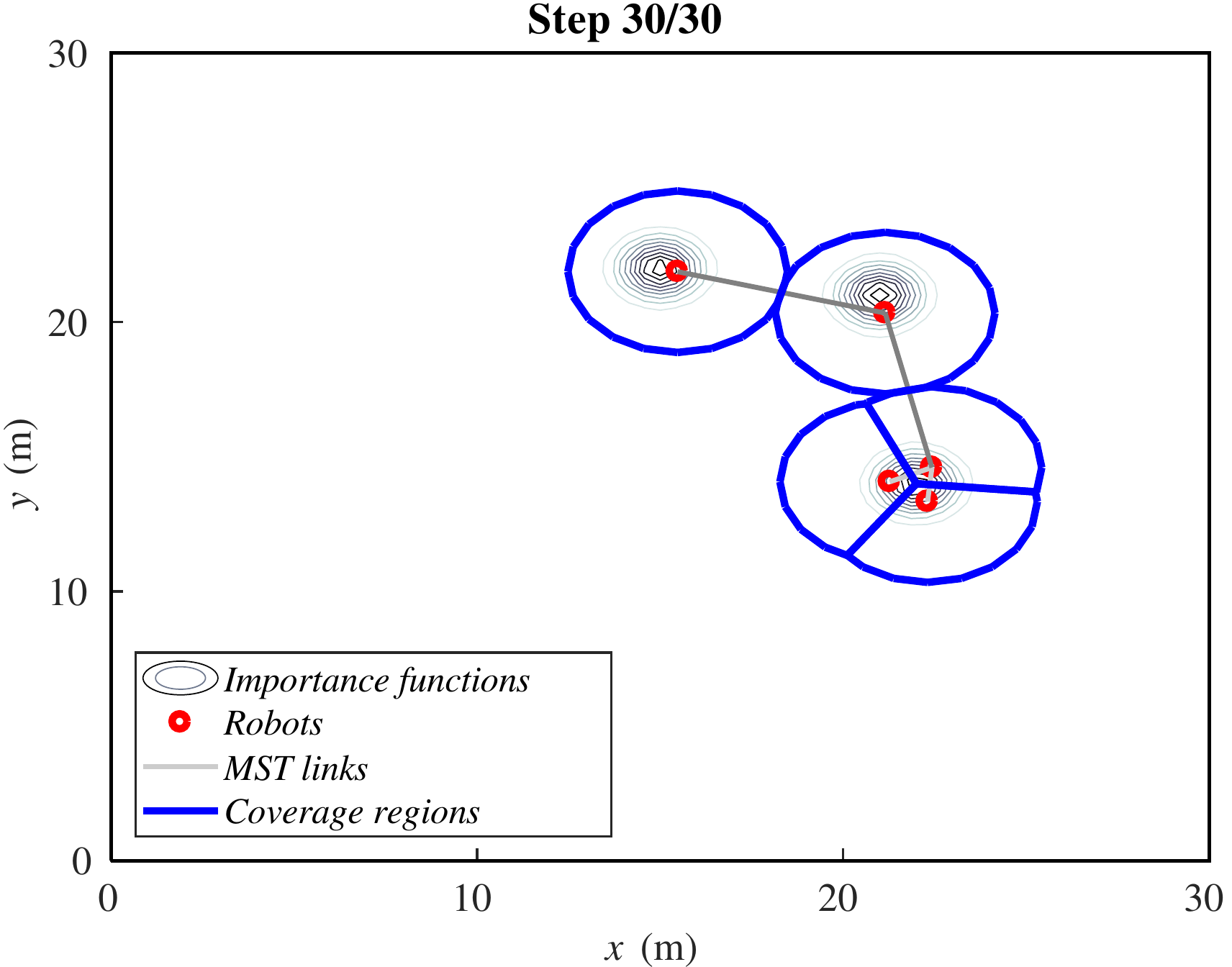}\\
	\end{tabular}
	
	\caption{Ground robots positions in four steps while covering three fixed targets which are assigned importance functions.}
	\label{progres_imp}
\end{figure}

Figure \ref{progres_imp} shows an example of the behavior of this method under fixed targets, in order to note how the agents deploy. Later, in Section \ref{simulations_section}, we will present more examples with moving targets. Agents are represented with small red circles and their communication tree is the set of gray lines that link them. The big blue circles represent the coverage regions of each agent. Finally, the grayscale concentric circles are the importance functions -with darker gray for higher values- where the targets are placed. In the first steps, the agents start moving towards the area of interest. When they reach it, they group around the first targets they get. Then they deploy to cover all of them.

The process executed at each iteration of the simulation is gathered in Algorithm \ref{importance_algorithm}.

\begin{algorithm}[H]
	\caption{Varying Importance Functions} 
	\label{importance_algorithm}
	\begin{doublespace}
		\begin{algorithmic}[1]
			\State Detect $(o_i)$
			\State Propagate $(o_i)$
			\State Update $(\phi (q))$
			\State Execute $(\text{Lloyd\_MSTConnCtrl \ Iteration})$ \Comment {Algorithm \ref{Lloyd_1step_connCtrl}}
		\end{algorithmic}
	\end{doublespace}
\end{algorithm}

We have observed that this method provides an accumulation of ground robots around the targets, which is not the behavior desired for coverage tasks. In addition, the local optimal behavior of this method sometimes implies that some targets may be uncovered. For these reasons, we propose another alternative, explained next.

\subsection{Fictitious Boundaries Redefining}
We propose another alternative based on redefining the boundaries of the working area. Since in coverage problems with constant density functions agents tend to evenly deploy over the area, here the idea is to modify in a virtual way the boundaries of this area to make agents adapt accordingly. The interest of this method lies in the even deployment of the agents over the modified working area. In addition, only the agents next to the virtual limits need to be informed about them. As far as we know, this solution has not been used before in the literature.

We decide to define the new area as the minimum rectangle including both targets and agents. As the area reduces, agents deploy with no different importances until the zone is as small as possible. If there are enough agents, they will spread so on to cover all the targets. So this method requires the aerial robots, that will act as leaders of the mission, to move in ways that can be covered by the ground robots. This restriction depends on some factors like the number of agents or their sensing radius, and will be studied in Section \ref{simulations_section}.

\begin{figure}[H]
	\centering
	\begin{tabular}{cc}
		\includegraphics*[width=0.45\textwidth]{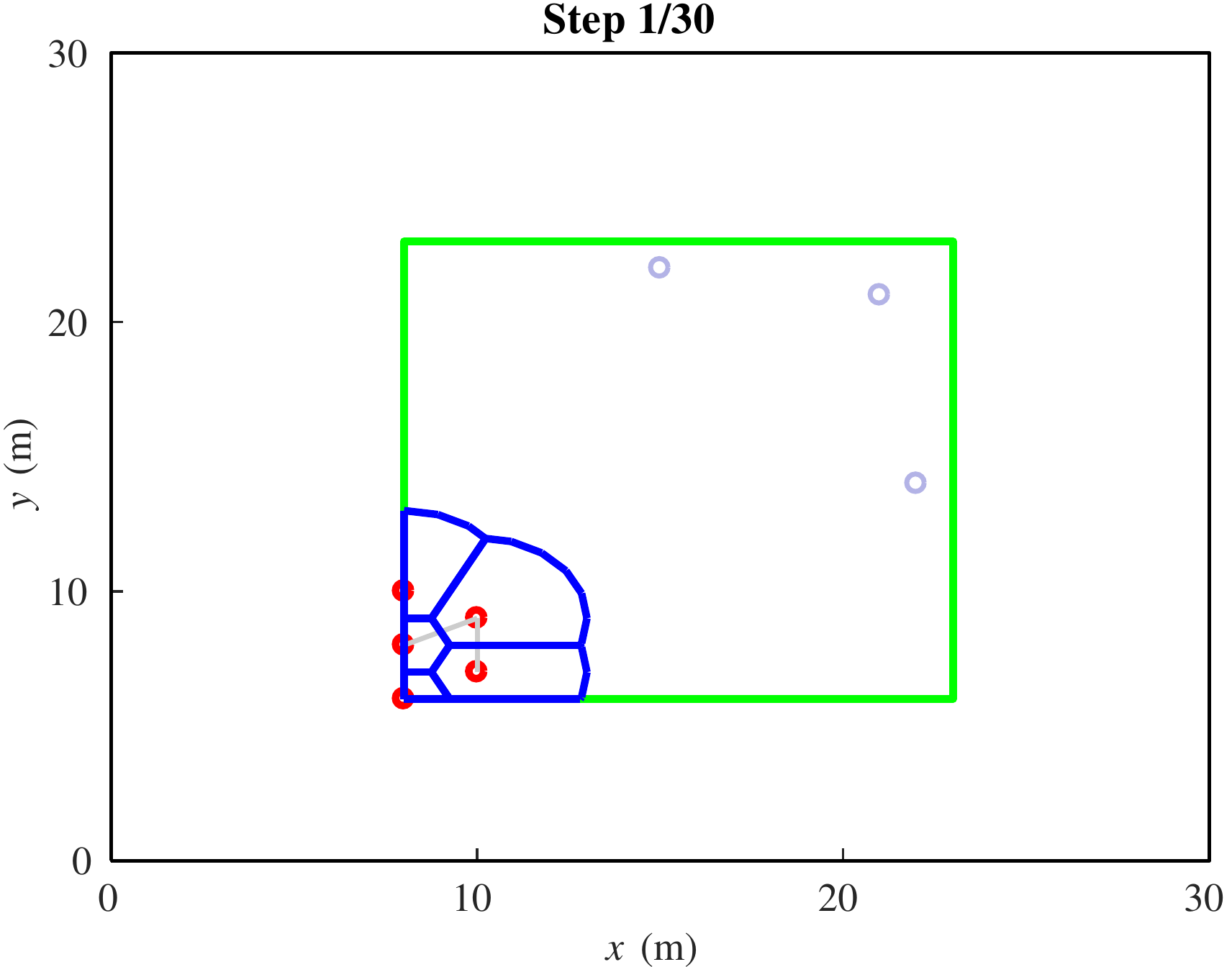}&
		\includegraphics*[width=0.45\textwidth]{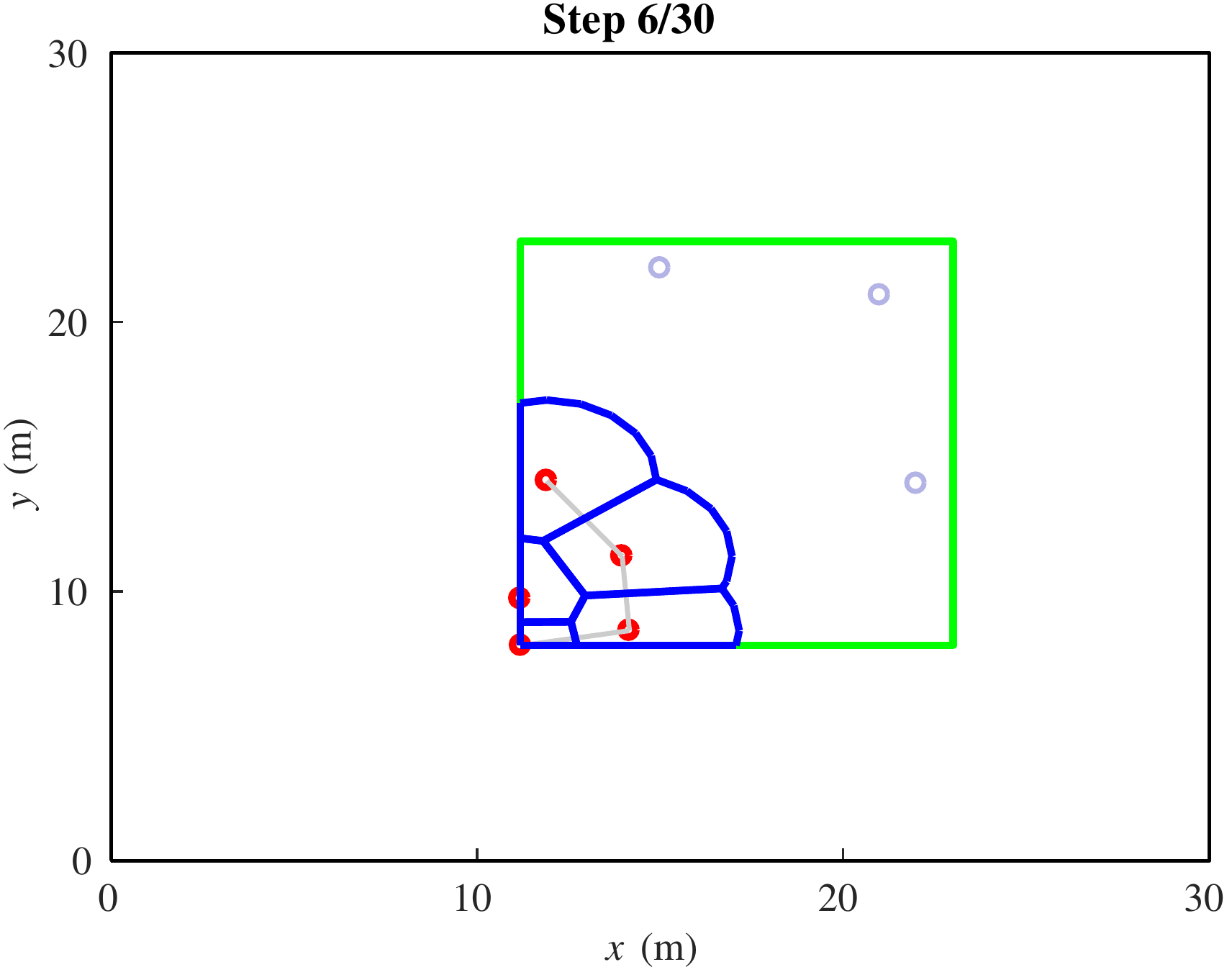}\\
		\includegraphics*[width=0.45\textwidth]{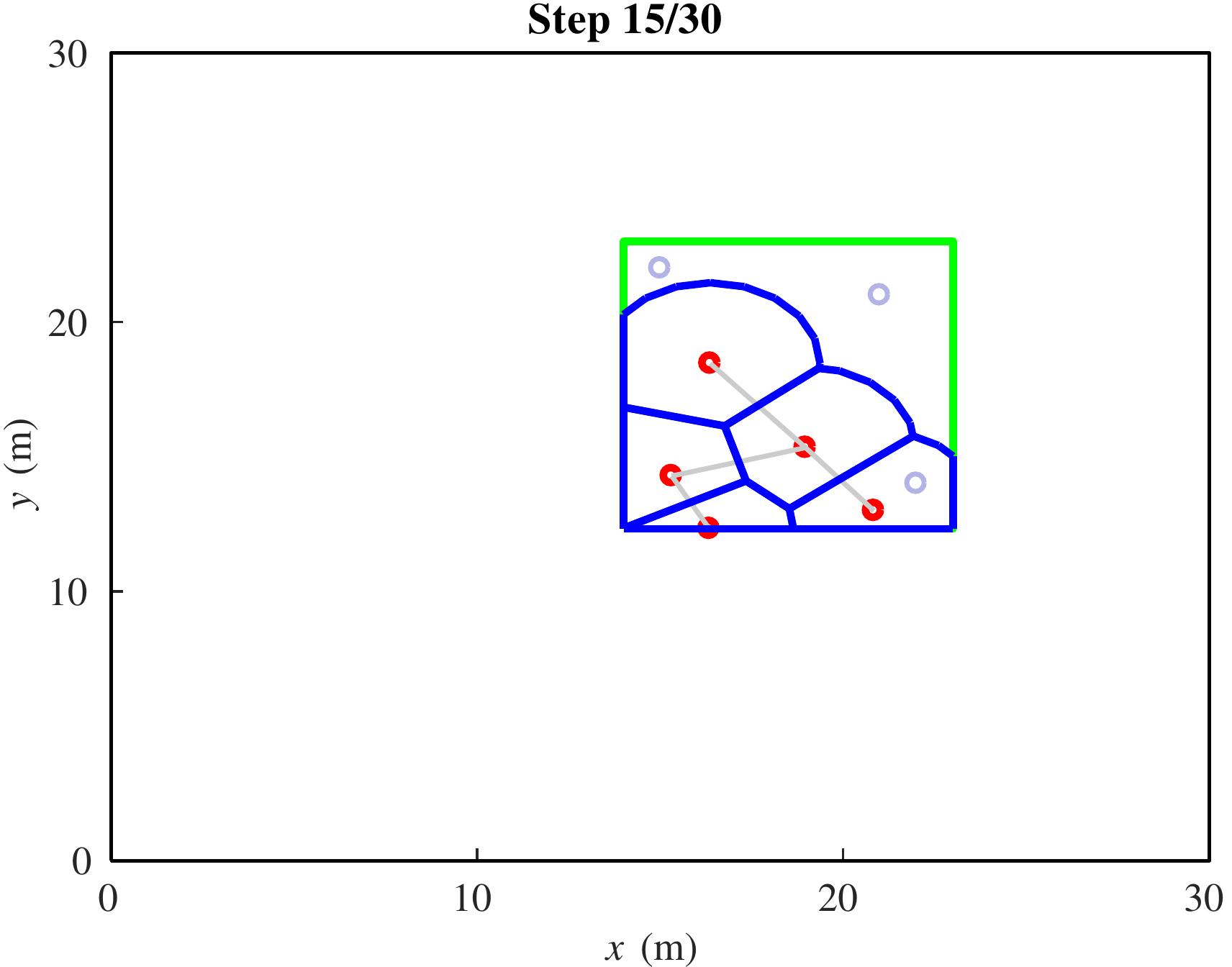}&
		\includegraphics*[width=0.45\textwidth]{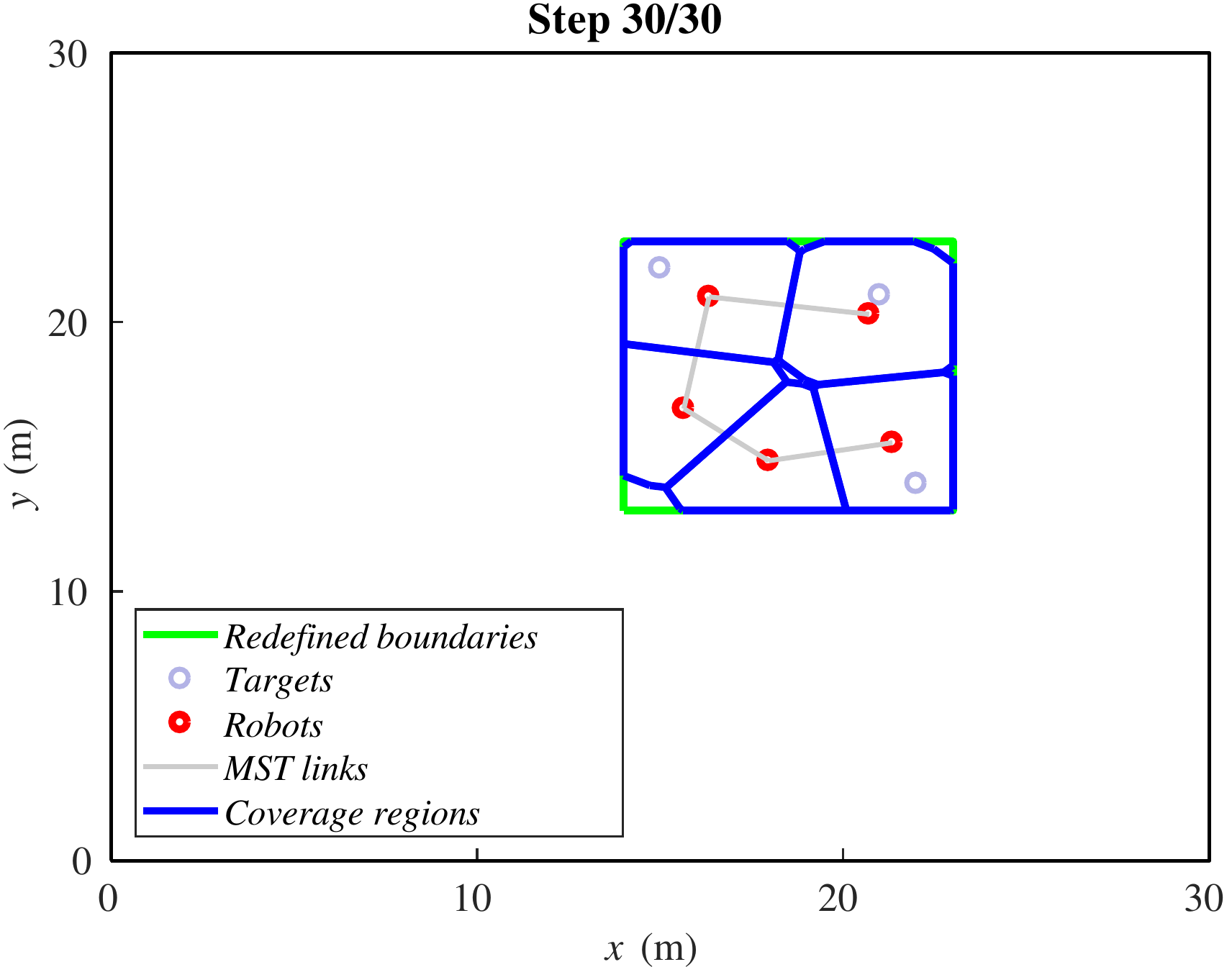}\\
	\end{tabular}
	
	\caption{Ground robots positions in four steps while covering three fixed targets delimited by fictitious boundaries.}
	\label{progres_reg}
\end{figure}

Figure \ref{progres_reg} shows the same example as the previous alternative. Here the redefined fictitious boundaries are the green lines and the targets are represented just with small gray circles. In first steps, the agents move towards the targets while the rectangle reduces. Once the rectangle is as small as targets allow, the agents spread over the final area. At the end, they cover a greater zone than the previous alternative, deploying uniformly. One can notice that the agents must have a fixed bearing reference in order to identify the orientation of the rectangle that iteratively defines the area to cover. In this paper, we make the assumption that they have this information. In real implementations, a mechanism for achieving this should be included. For instance, agents can be equipped with a compass, or run a localization method, as in \cite{Stegagno-Tro2016}.

The process that this alternative runs in a centralized way at each simulation step to follow the targets is shown in Algorithm \ref{boundaries_algorithm}.

\begin{algorithm}[H]
	\caption{Fictitious Boundaries Redefining} 
	\label{boundaries_algorithm}
	\begin{doublespace}
		\begin{algorithmic}[1]
			\State Detect $(o_i)$
			\State $x_{min,Area} \ = \ min(p_{i,x},o_{i,x})$
			\State $x_{max,Area} \ = \ max(p_{i,x},o_{i,x})$
			\State $y_{min,Area} \ = \ min(p_{i,y},o_{i,y})$
			\State $y_{max,Area} \ = \ max(p_{i,y},o_{i,y})$
			\State Area $=[x_{min,Area},x_{max,Area},y_{min,Area},y_{max,Area}]$
			\State $Q=$ Area
			\State Execute $(\text{Lloyd\_MSTConnCtrl \ Iteration})$ \Comment {Algorithm \ref{Lloyd_1step_connCtrl}}
		\end{algorithmic}
	\end{doublespace}
\end{algorithm}

\subsection{Comparison between Both Alternatives}
The example studied allows making a comparison between the proposed alternatives. In addition to the previously discussed differences between the amount of agents that need to be informed of the positions of the moving aerial robots, in this section we explain some behavioral differences.

On the one hand, the importance functions alternative tend to group the ground robots around the targets. Depending on the initial relative positions, this could make leave some others uncovered. On the other hand, the method redefining the boundaries achieves a wider formation, covering a greater zone than the first alternative. So this method focuses on the area around the targets, while the first method focus on the targets themselves.

Considering this information, the importance functions method is better suited for individual objectives whose environment is not important, while the boundaries redefining alternative is more appropriate for concentrations of targets or wide areas of interest.

\section{Simulations and Results}
\label{simulations_section}

Both alternatives presented so far, are built on deployment methods which were originally designed for static setups. Since our aim is to use these methods to dynamically track variations in the region to be covered, a question that arises is how slow the modifications in the environment must be in order for the agents to efficiently track them. This question is hard to answer from an analytical point of view. Moreover, this speed may depend on several facts (number of agents, area to be covered, sensing radius, communication radius, etc.). In this section we make a thorough parametric study to identify the factors that influence the allowable speed. Here, the simulations are only executed with MATLAB in order to get repeatable results, abstracting from implementation issues that carry 3D simulation or real experiments. Later, in Sections  \ref{realistic_sim_section} and \ref{experiment_section}, we carry out 3D simulations and experiments that involve other specific problems (robot kinematics, collision avoidance, noisy data, or restrictions on the motions).

We define now the base experiment for the parametric study. We have a group of six ground robots implementing the boundaries redefining algorithm. We will consider the particular case where the targets fly in a rectangular formation composed of twelve aerial robots, and move 0.3 meters per simulation step. The sensing radius of the ground robots is $s=3$ m, and their communication radius is $r=6$ m.

Starting from this experiment, we are varying four parameters in order to study how they influence in the algorithm behavior. These four parameters are the velocity of the formation to track, the sensing radius, the number of agents, and the tracking method.

\subsection{Formation Velocity}
\begin{figure}[H]
	\centering
	\includegraphics[width=0.45\textwidth]{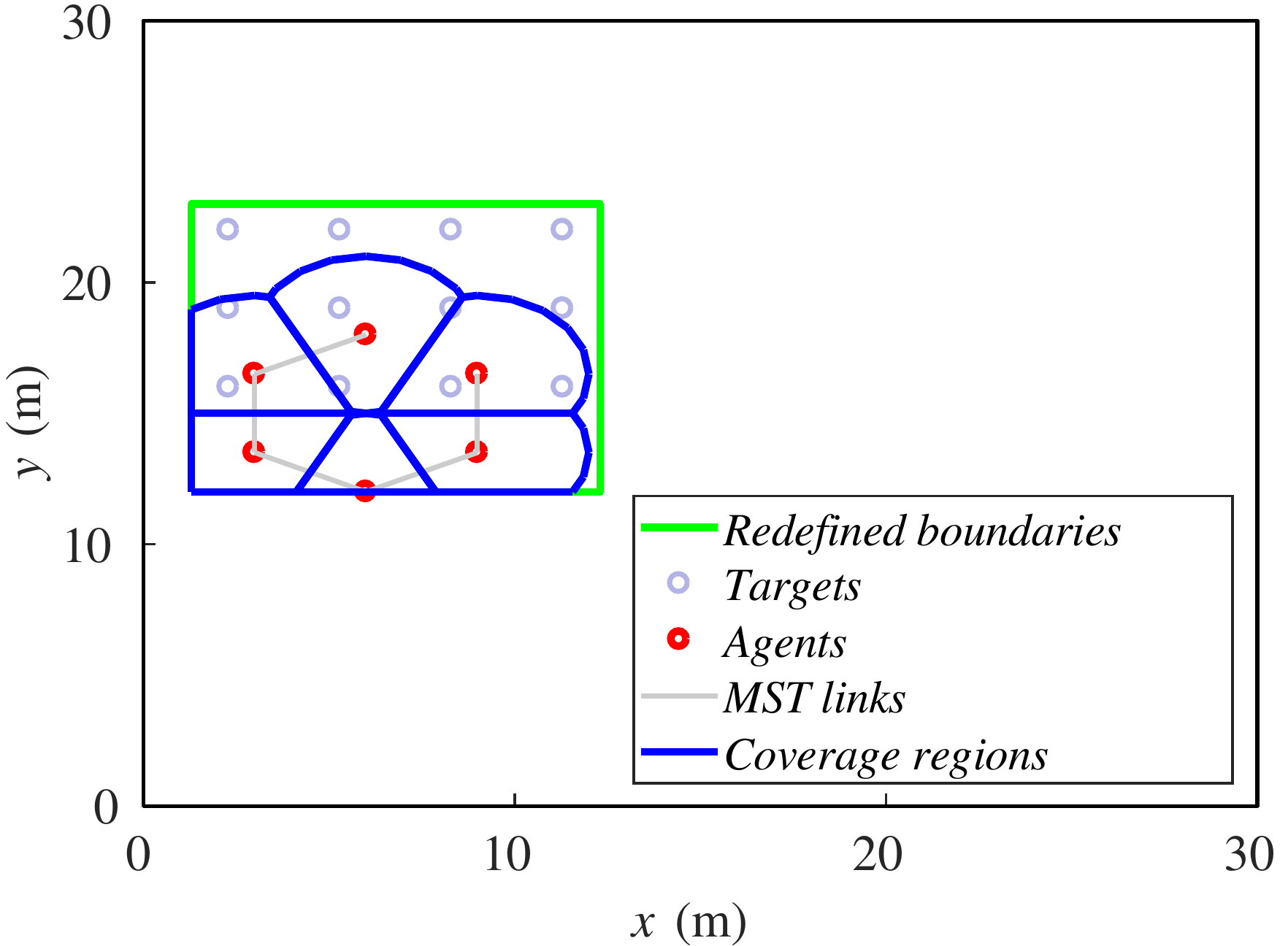}
	\caption{Initial positions of agents and targets.}
	\label{init_position}
\end{figure}

In the first study we vary the formation velocity in the interval 0.25--0.5 m/step. The formation of targets moves from left to right and its initial position is described in Figure \ref{init_position}, as well as the initial position of the ground agents.

\begin{figure}[H]
	\centering
	\begin{tabular}{cc}
		\includegraphics*[width=0.49\textwidth]{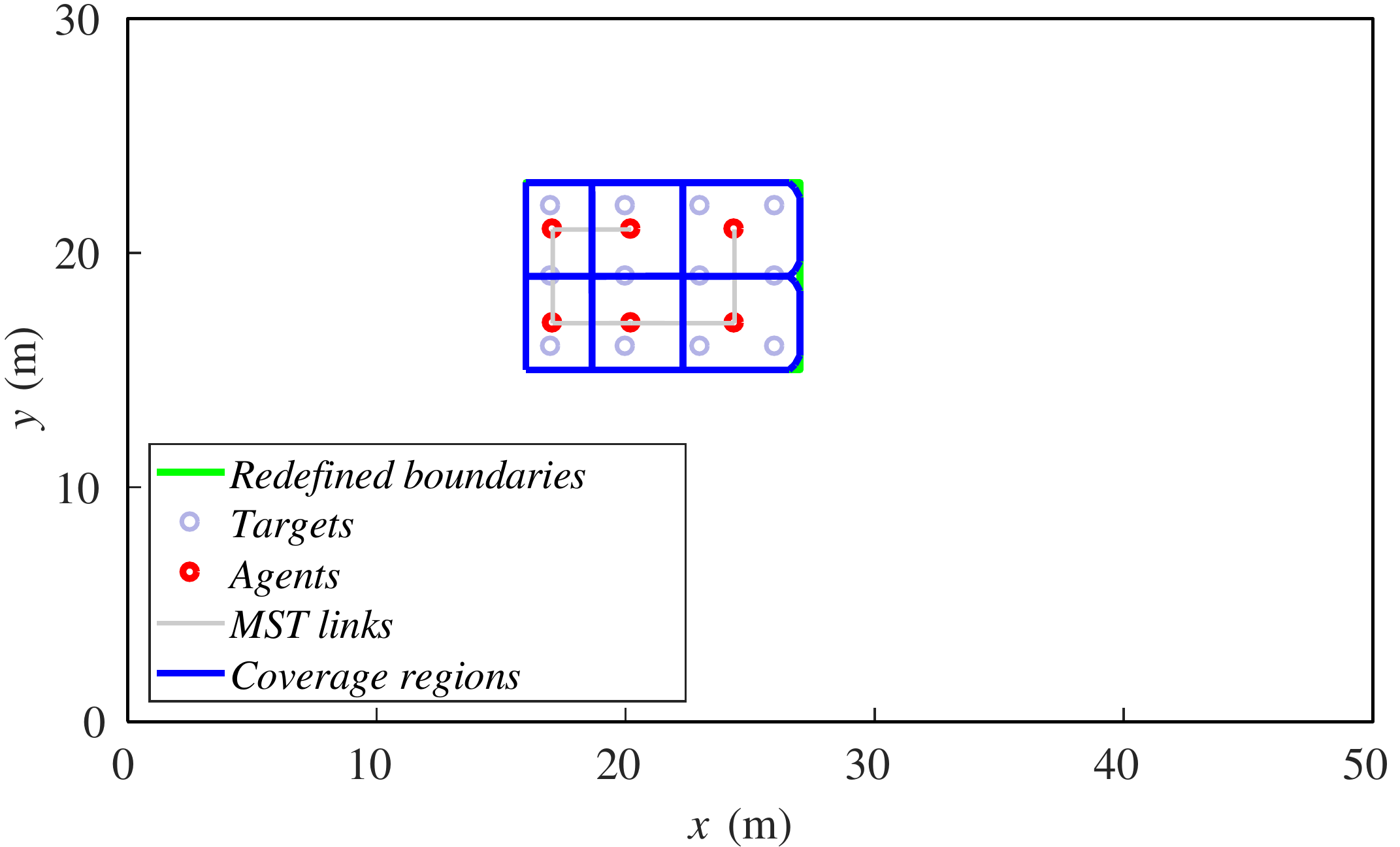}&
		\includegraphics*[width=0.49\textwidth]{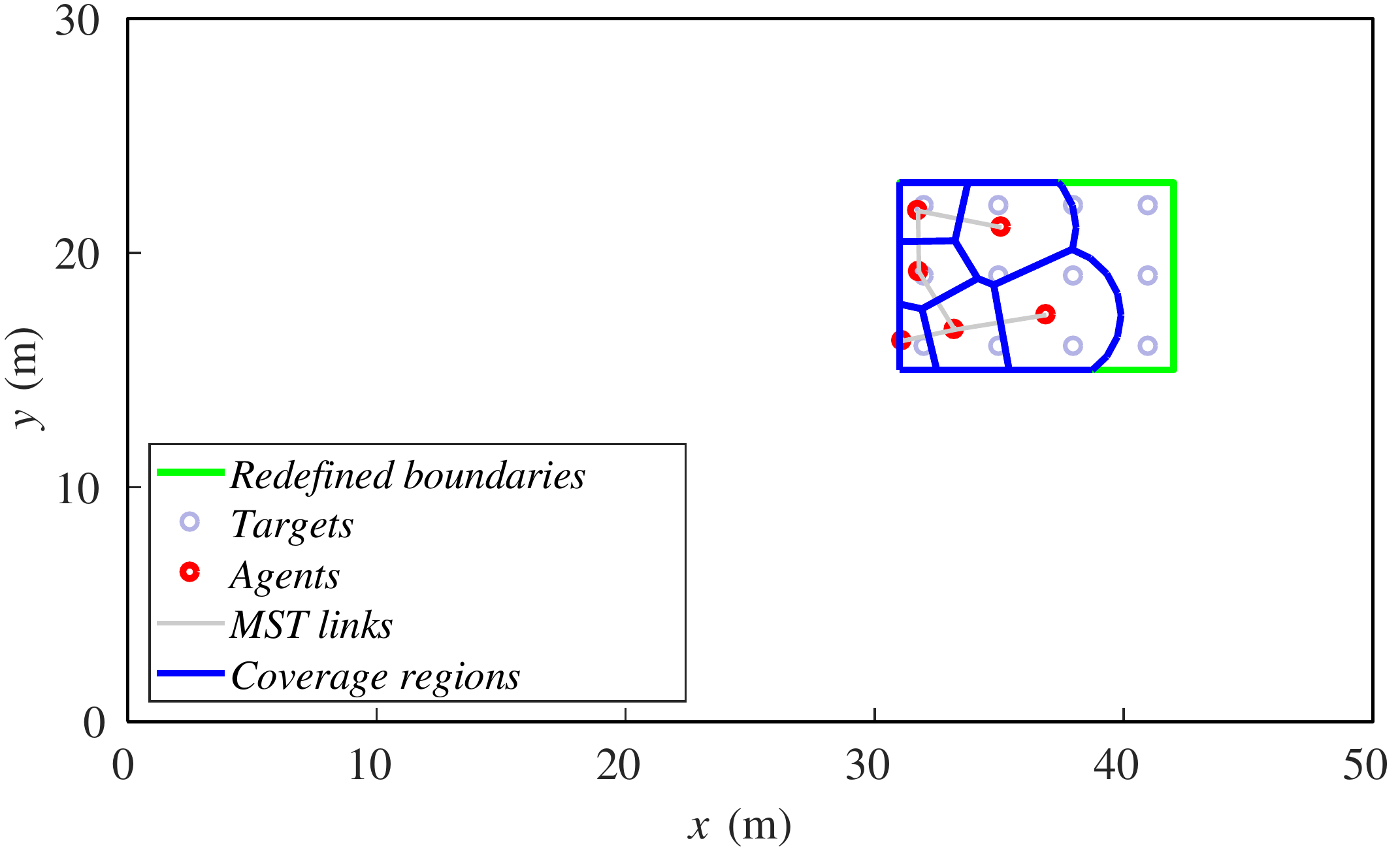}\\		
		{\footnotesize \textbf{(a)}} & {\footnotesize \textbf{(b)}}
	\end{tabular}	
	\caption{Final positions of agents and targets for (\textbf{a}) 0.25 m/step and (\textbf{b}) 0.5 m/step.}
	\label{final_pos_vel}
\end{figure}

Since the velocity is different on each case and in all of the setups we run 60 steps of the method, the final position of the formation will be different. In Figure \ref{final_pos_vel} we show the final dispositions of all the robots for 0.25 and 0.5 m/step cases. For lower velocities, the algorithm is able to keep the formation covered as it moves. However, if velocity increases the agents can not correctly cover the whole formation. Some of them agglomerate in the rear formation leaving the front targets uncovered.

\begin{figure}[H]
	\centering
	\begin{tabular}{cc}
		\includegraphics*[width=0.45\textwidth]{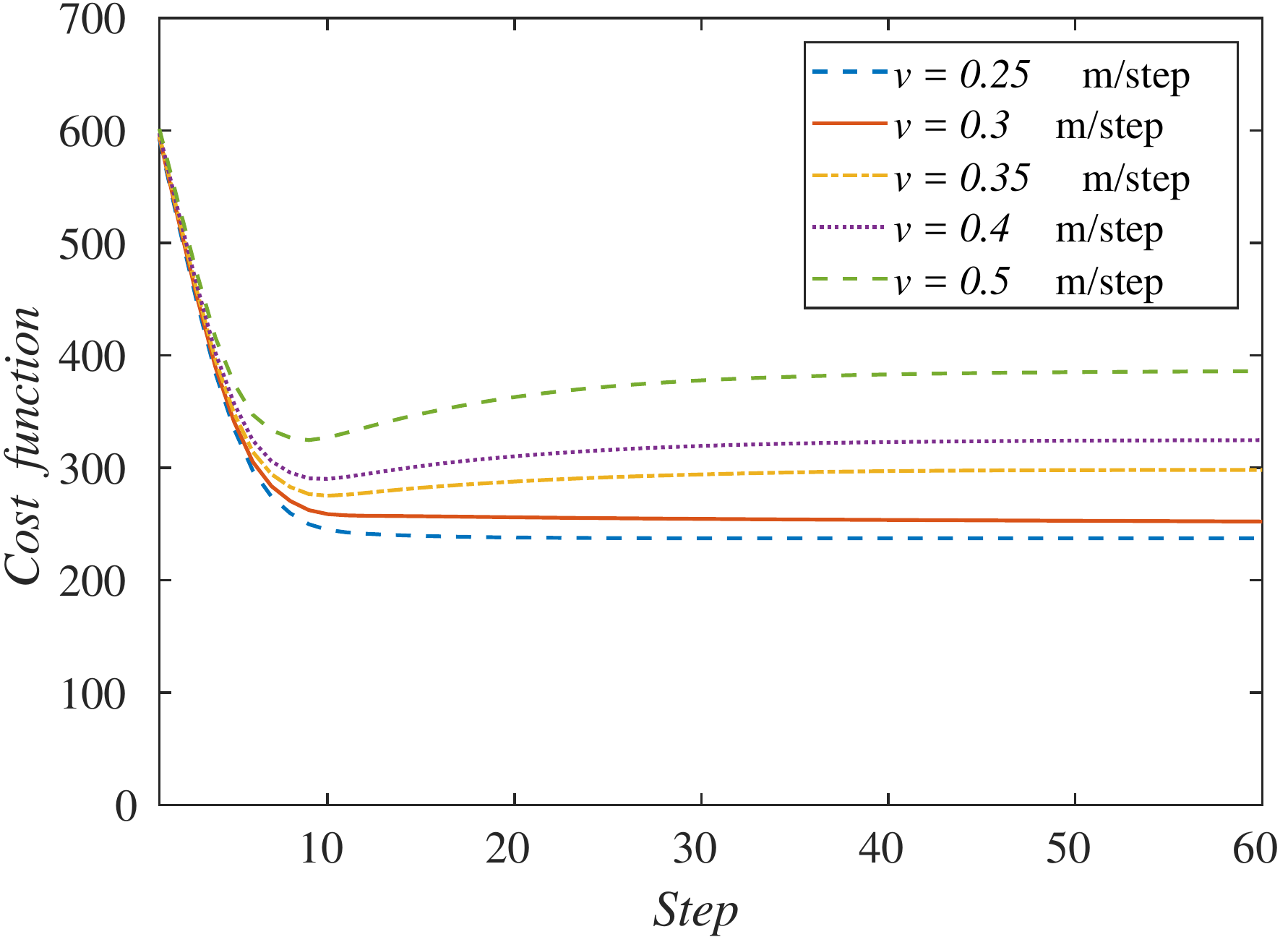}&
		\includegraphics*[width=0.45\textwidth]{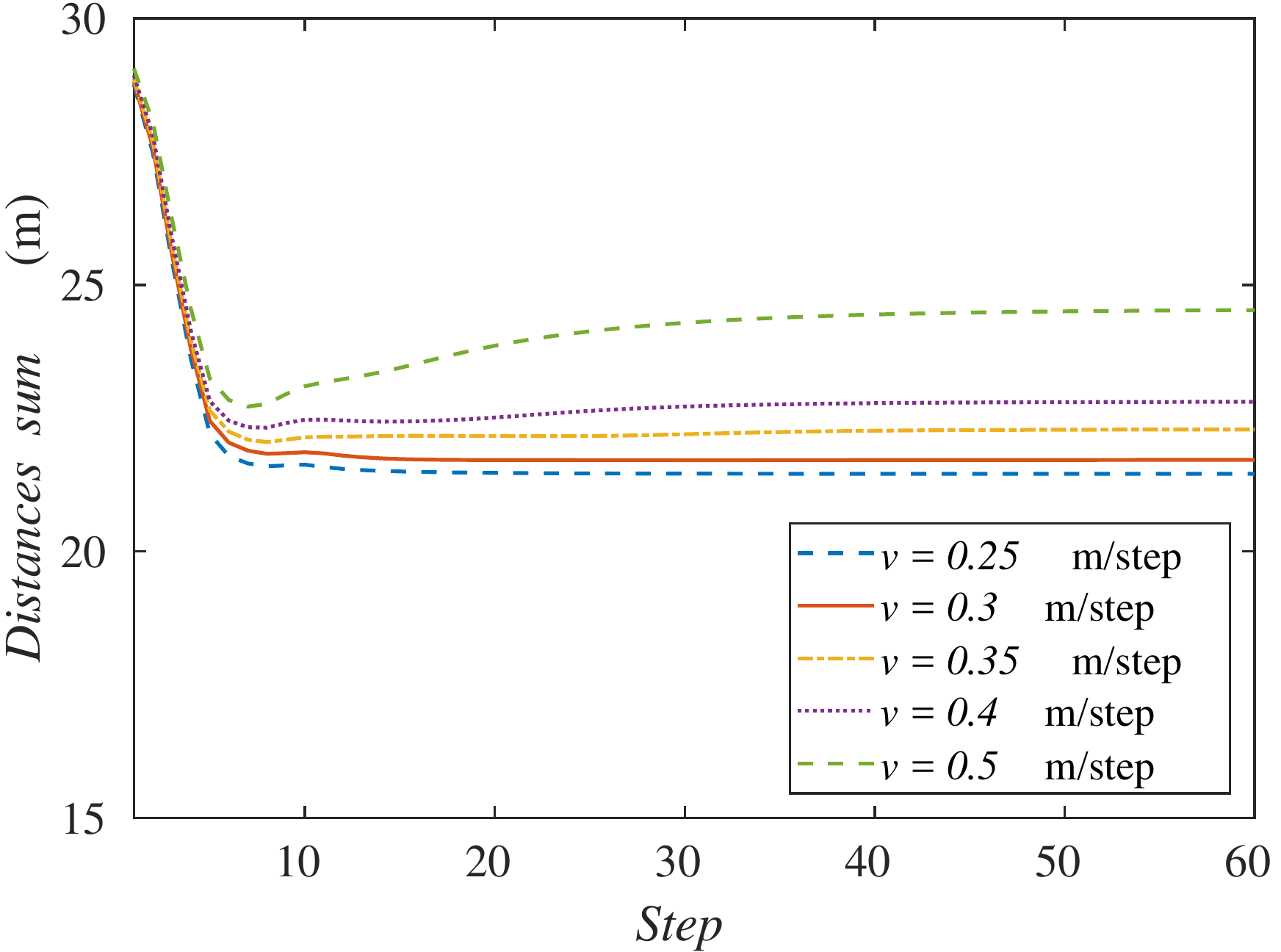}\\
		{\footnotesize \textbf{(a)}} & {\footnotesize \textbf{(b)}}
	\end{tabular}	
	\caption{Performance metrics, (\textbf{a}) cost function and (\textbf{b}) sum of agents distances to the formation center, along the simulation depending on the formation velocity.}
	\label{metrics_vel}
\end{figure}

Now in Figure \ref{metrics_vel}(a) we represent the cost functions (Eq.~\ref{cost_function}) for each velocity along the simulation steps. One can realize that the cost function stabilizes in a constant value for each velocity. This behavior means that in all cases the agents spread to a constant disposition and reach the speed of the formation, advancing with it. Furthermore, the metric minimizes for the lowest speeds, because the agents cover the whole formation of the targets, as we have seen in Figure \ref{final_pos_vel}.

The behavior of the algorithm can also be monitored with a variable defined as 

\begin{equation}
\label{robot_dist}
\sum_{i=1}^n \left\| p_i-O \right\|,
\end{equation}

\noindent i.e., the sum of the distances of each ground robot to the center of the formation, $O$. This variable can describe the quality of the coverage performed. Figure \ref{metrics_vel}(b) gathers this sum of distances along the simulation with the velocities studied. We observe the same behavior of the curves as in the cost function. In case of higher velocities the sum of distances increases since the agents can not cover the whole formation. The metric minimizes also for lower velocities. For this reason, from now on we will show this performance metric (Eq.~\ref{robot_dist}), since it is easier to understand and its behavior keeps similar to the cost function in the experiments remaining.

The results obtained evidence the existence of a maximum formation speed that the algorithm can correctly track. In \cite{Martinez-TAC07b} authors prove that the static coverage task for an one-dimensional case, with $\phi (q)=1$ and an relative precision $\varepsilon$, is achieved in a time that polynomially depends on the number of agents. Our case is dynamic and two-dimensional, so this study would imply a much higher complexity. We obtain the maximum speed experimentally instead. Taking a look at the graphs, we conclude that this maximum velocity is around 0.3 m/step for the described base setup. That is why the base experiment is designed with this formation velocity, in order to highlight the differences in the following studies.

Note that the velocity units are m/step. This will allow a higher maximum speed of the targets for systems that execute quicker a step of the algorithm. This includes, e.g., the communication between the agents, the computational load or the motion restrictions of the ground robots.

\subsection{Sensing Radius}
In this case we test five scenarios with sensing radius in the interval 2--4 meters. In order to maintain the relation $s=r/2$ discussed in Section \ref{coverage_section}, the communication radius is changed proportionally in the simulations. Initial positions are the same as in the previous experiments and the simulations are run with 80 steps so that every experiment reaches the steady state.

\begin{figure}[!b]
	\centering
	\begin{tabular}{cc}
		\includegraphics*[width=0.45\textwidth]{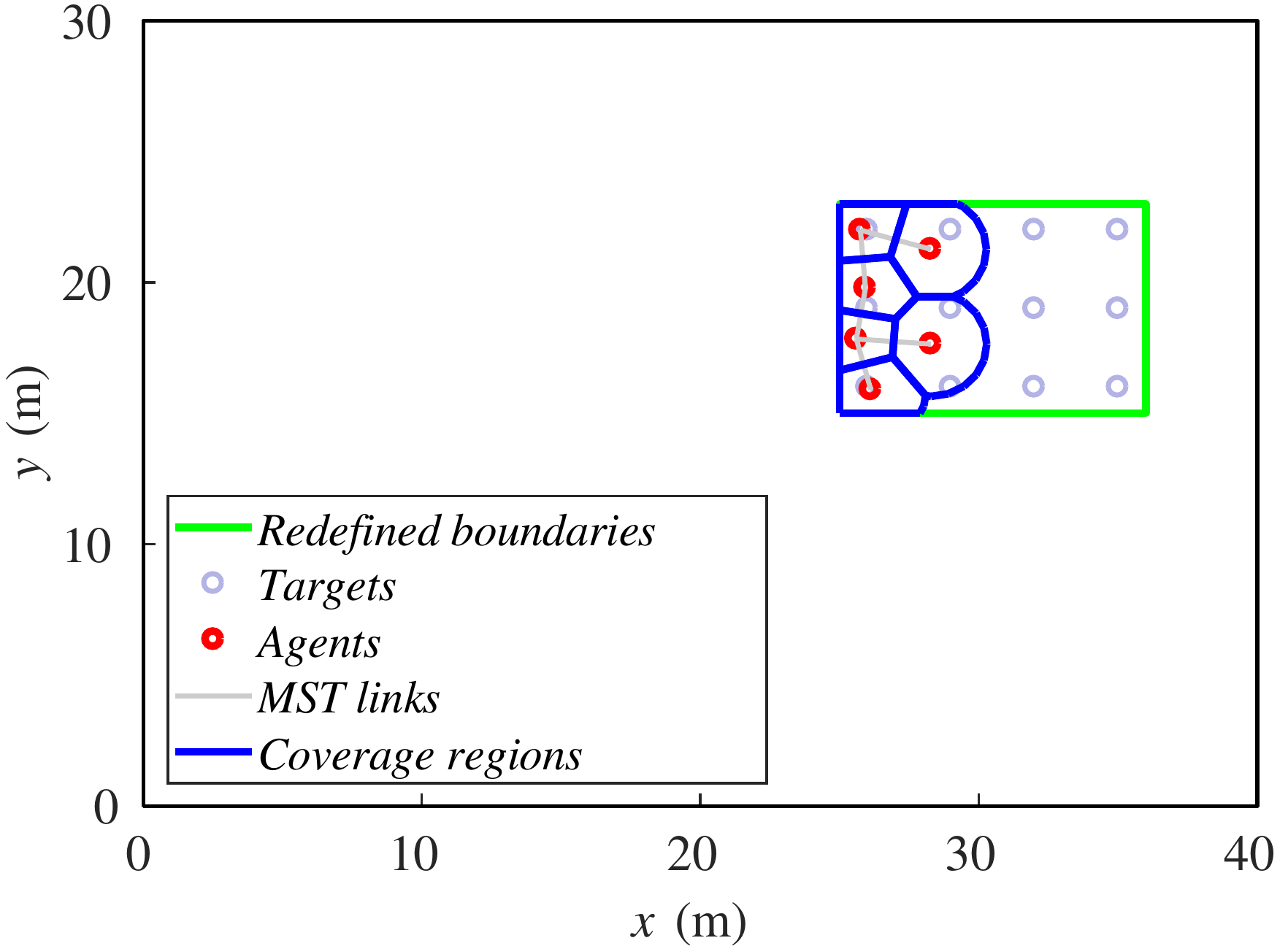}&
		\includegraphics*[width=0.45\textwidth]{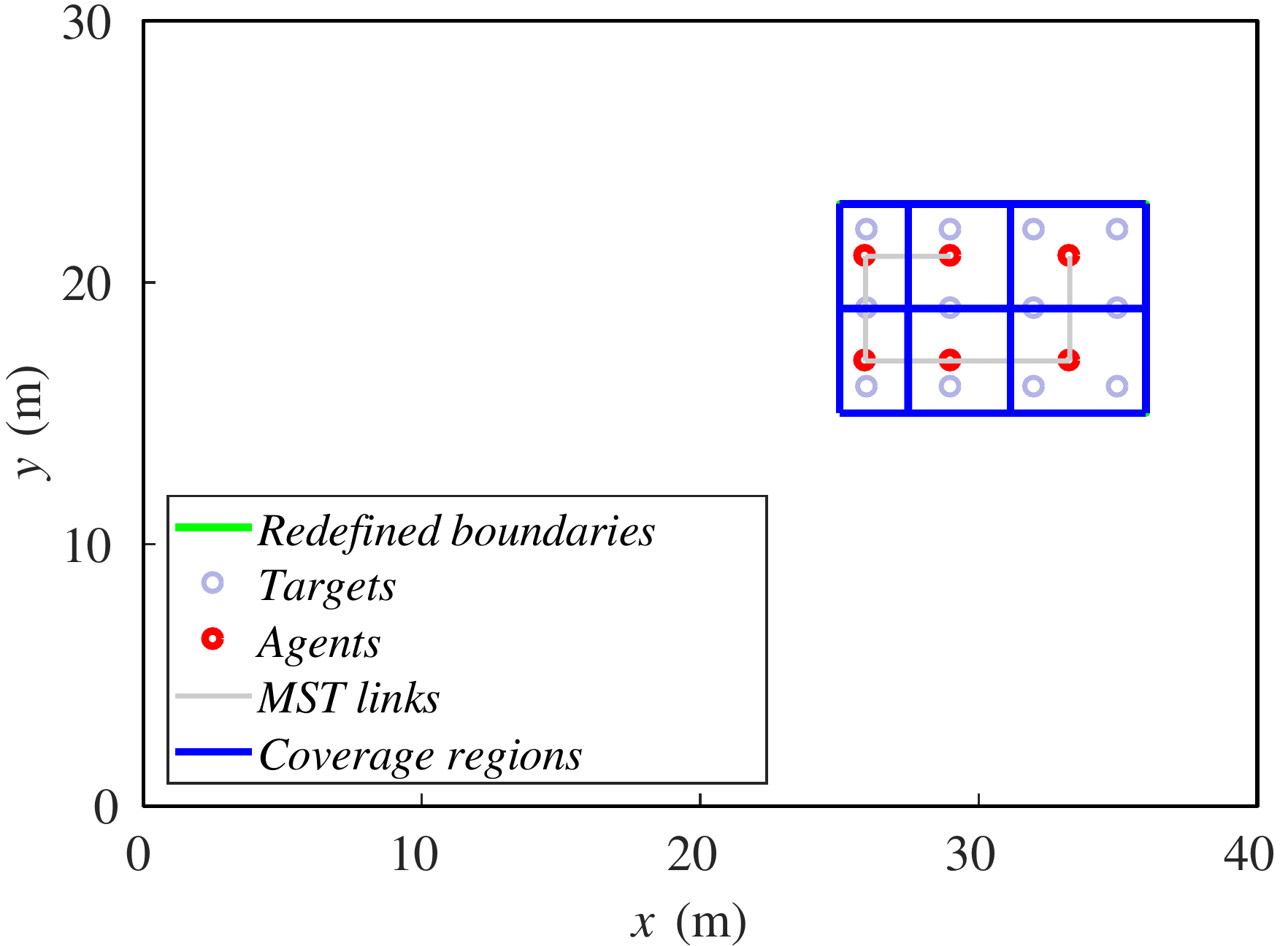}\\
		{\footnotesize \textbf{(a)}} & {\footnotesize \textbf{(b)}}
	\end{tabular}	
	\caption{Final positions of agents and targets for (\textbf{a}) 2 m and (\textbf{b}) 3.5 m.}
	\label{final_pos_rad}
\end{figure}

The sensing radius is often one of the parameters that affect the allowable speed of the formation to be tracked. Showing the final positions of the robots for 2 and 3.5 meters (Figure \ref{final_pos_rad}), one can conclude that, with lower sensing radius, the agents can not perform a wide deployment to cover the targets, and they accumulate at the rear formation.

\begin{figure}[H]
	\centering
	\includegraphics[width=0.5\textwidth]{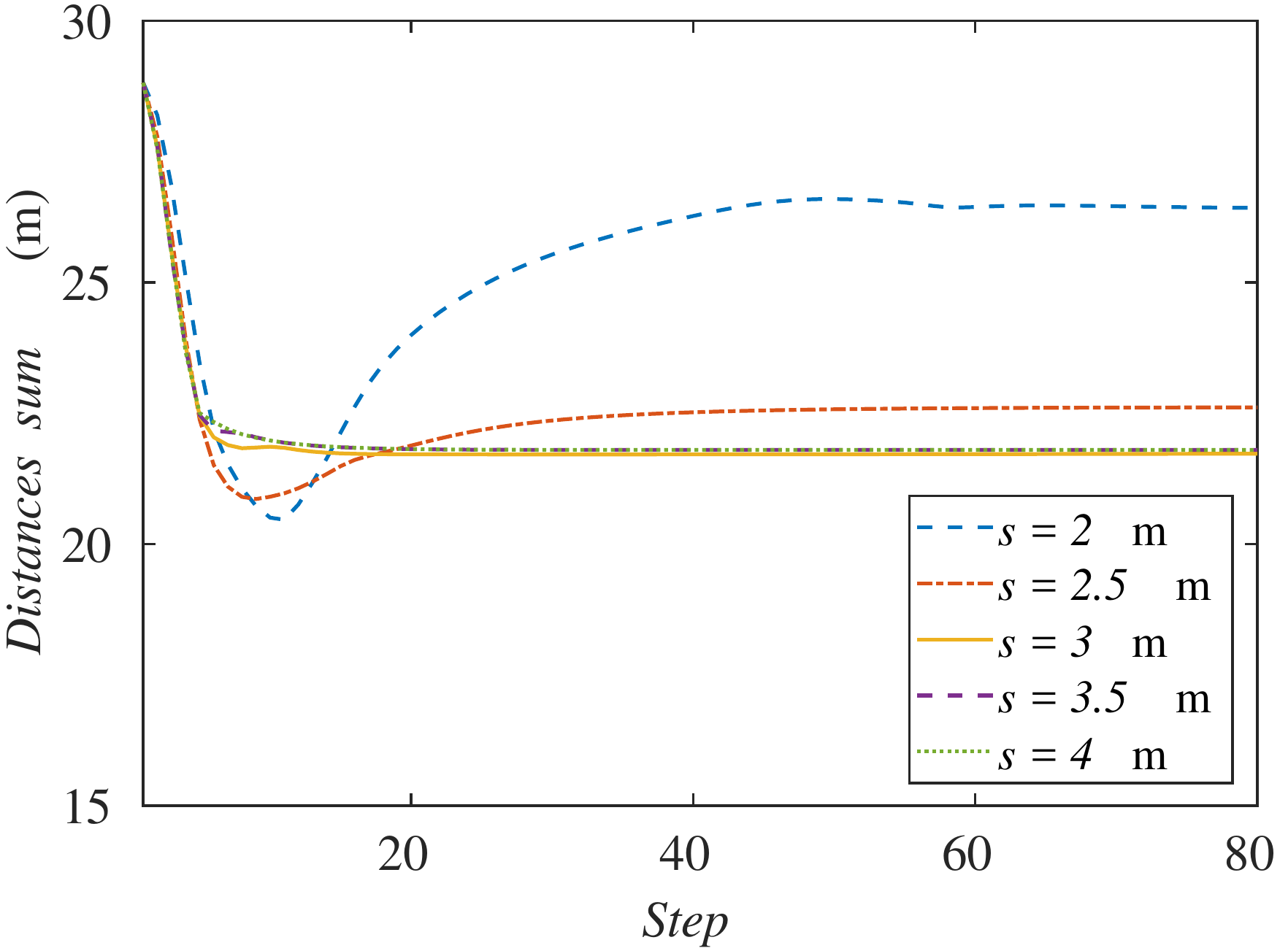}
	\caption{Sum of agents distances to the formation center along the simulation depending on the sensing radius.}
	\label{robotdist_rad}
\end{figure}

Figure \ref{robotdist_rad} represents the sum of the distances from the ground agents to the formation center for each sensing radius. As we said before, the smaller radius clearly make this variable increase because the agents get to the back of the formation. On the other hand, while the radius increases we can observe that the curves tend to the same value. This behavior is due to the fact that the agents have an equilibrium distribution that balances their advance with the formation speed. Even though the radius were much higher, the balance position would be the same.

From the results, we can observe that as the sensing radius of agents decreases, they have more difficulties to follow the moving formation. One of the possible reasons for this behaviors is that, at every step, agents perform motions that keep them inside the intersection between their own Voronoi and sensing region. When this region is smaller, agents perform motions that may be too short and may avoid them tracking the aerial formation.

\begin{figure}[!b]
	\centering
	\includegraphics[width=0.5\textwidth]{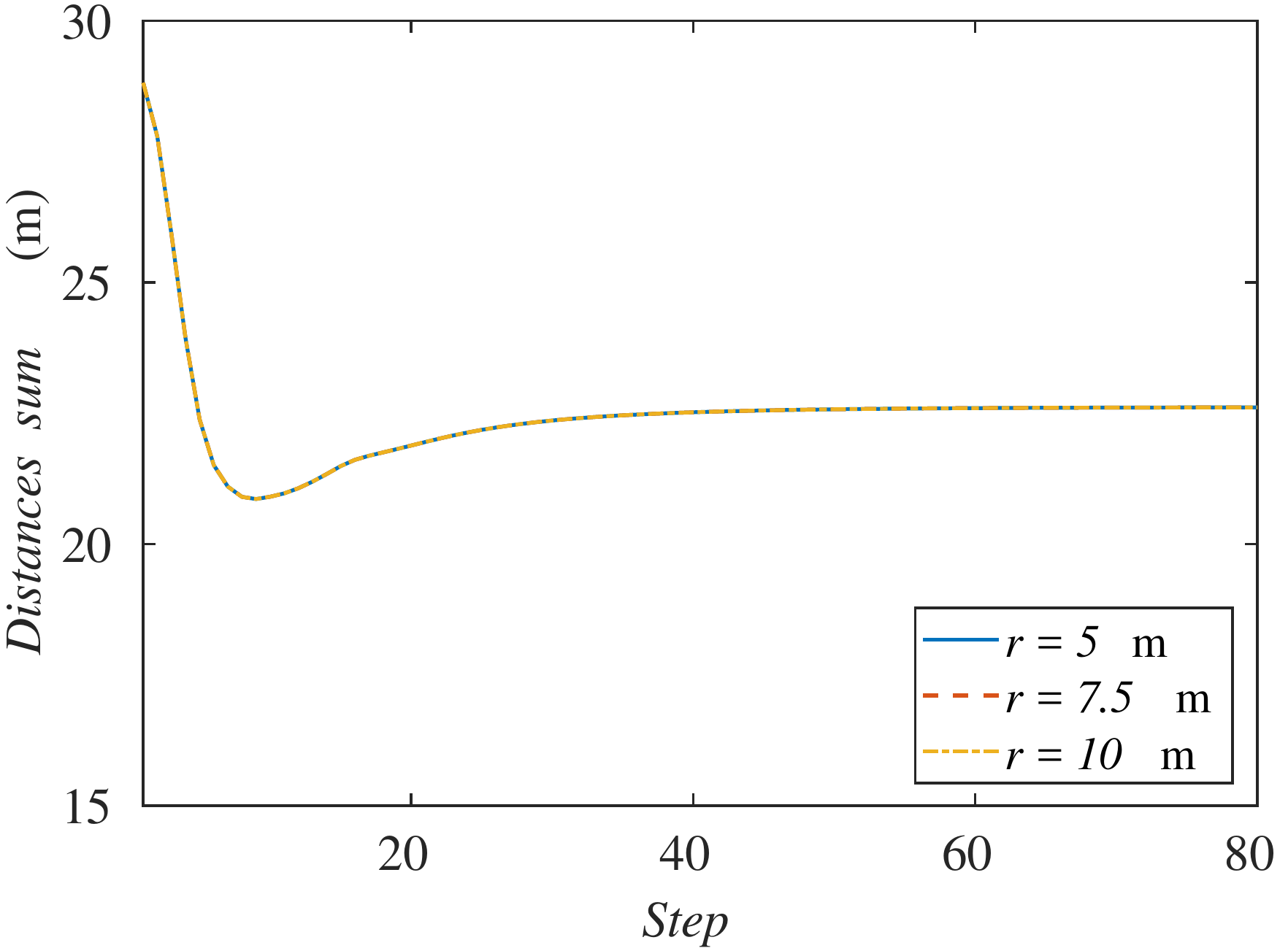}
	\caption{Sum of agents distances to the formation center along the simulation depending on the communication radius.}
	\label{robotdist_comm_rad}
\end{figure}

Another possible explanation would be that the communication links were too restrictive for the motion of the agents. Recall that with each variation of the sensing radius we have changed the communication radius too in order to keep the relation $r=2s$. If the communication radius were higher, one could think that the agents may expand over the region and cover all the targets. For this reason, we have run another set of simulations with larger communication radius, keeping a sensing radius of 2.5 m, that in the previous experiment could not cover all the formation. In Figure \ref{robotdist_comm_rad}, we can see the results of the sum of distances to the center of the formation varying the communication radius from 5 to 10 meters. We can observe that all the graphs for each communication radius concur in the same curve. So we can conclude that this is not the reason why the agents can not follow the formation with lower sensing radius.

These results also let us think that the connectivity maintenance method is not interfering with the coverage and tracking goal, as far as the region can be indeed covered by the agents.

\subsection{Number of Agents}
This time we vary the number of ground robots in the system, duplicating them iteratively in order to highlight the differences in the experiments. In this case we test from 4 to 16 agents. The initial positions can not be the same because the number of ground robots is variable. The agents start from positions that are close to the previous experiments, and run a 300 step simulation in order to be able to observe the behavior of the curve with the highest number of agents.

\begin{figure}[H]
	\centering
	\includegraphics[width=0.5\textwidth]{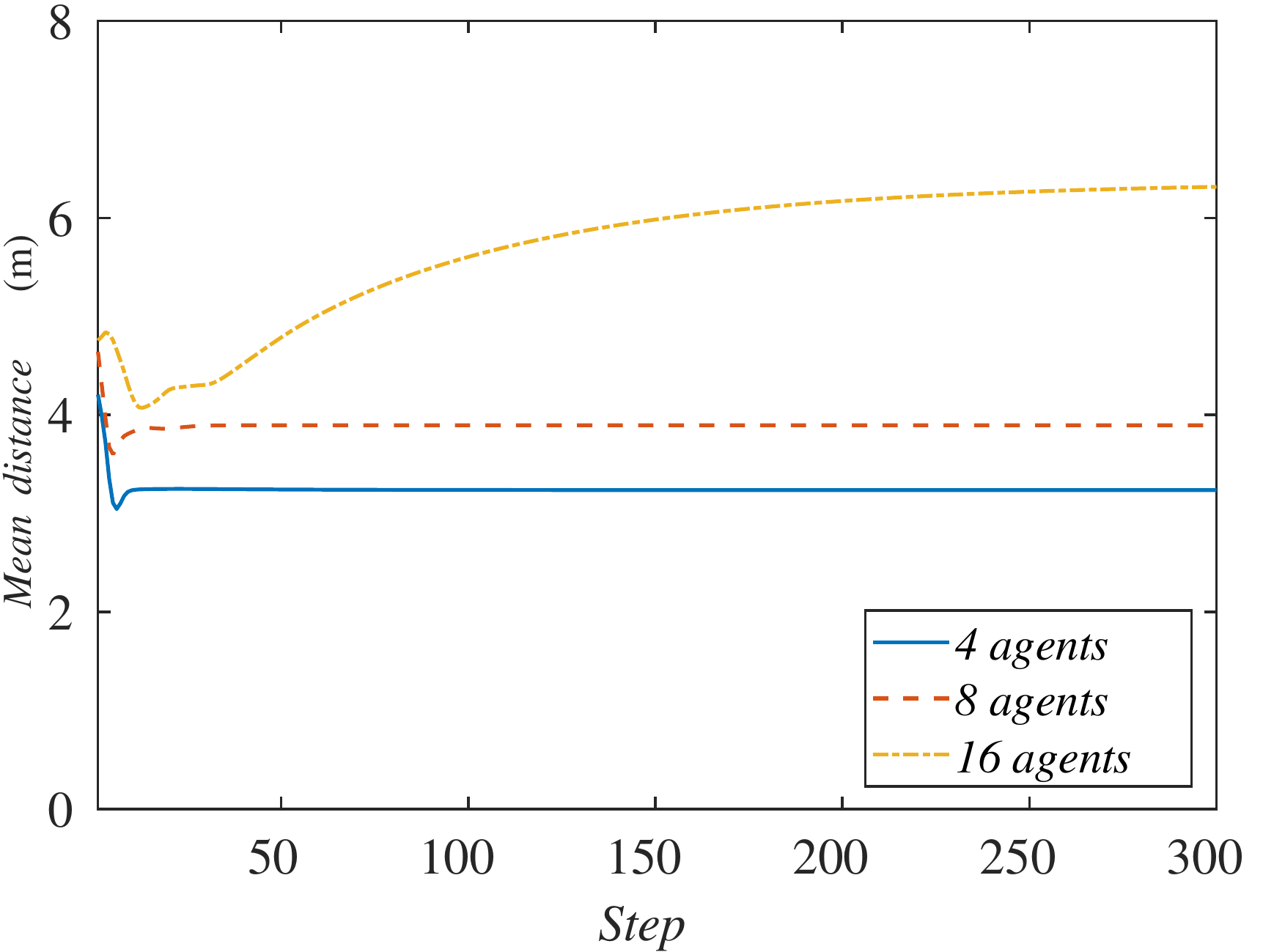}
	\caption{Mean agents distance to the formation center along the simulation depending on the number of agents.}
	\label{robotdist_rob}
\end{figure}

Due to the fact that the number of agents is variable, now the sum of the agents distances to the formation center is not an appropriate variable to represent. We graph the mean distance instead in Figure \ref{robotdist_rob}. The curves do not start from the same point as in the previous experiments because of this same reason.

\begin{figure}[!t]
	\centering
	\begin{tabular}{cc}
		\includegraphics*[width=0.45\textwidth]{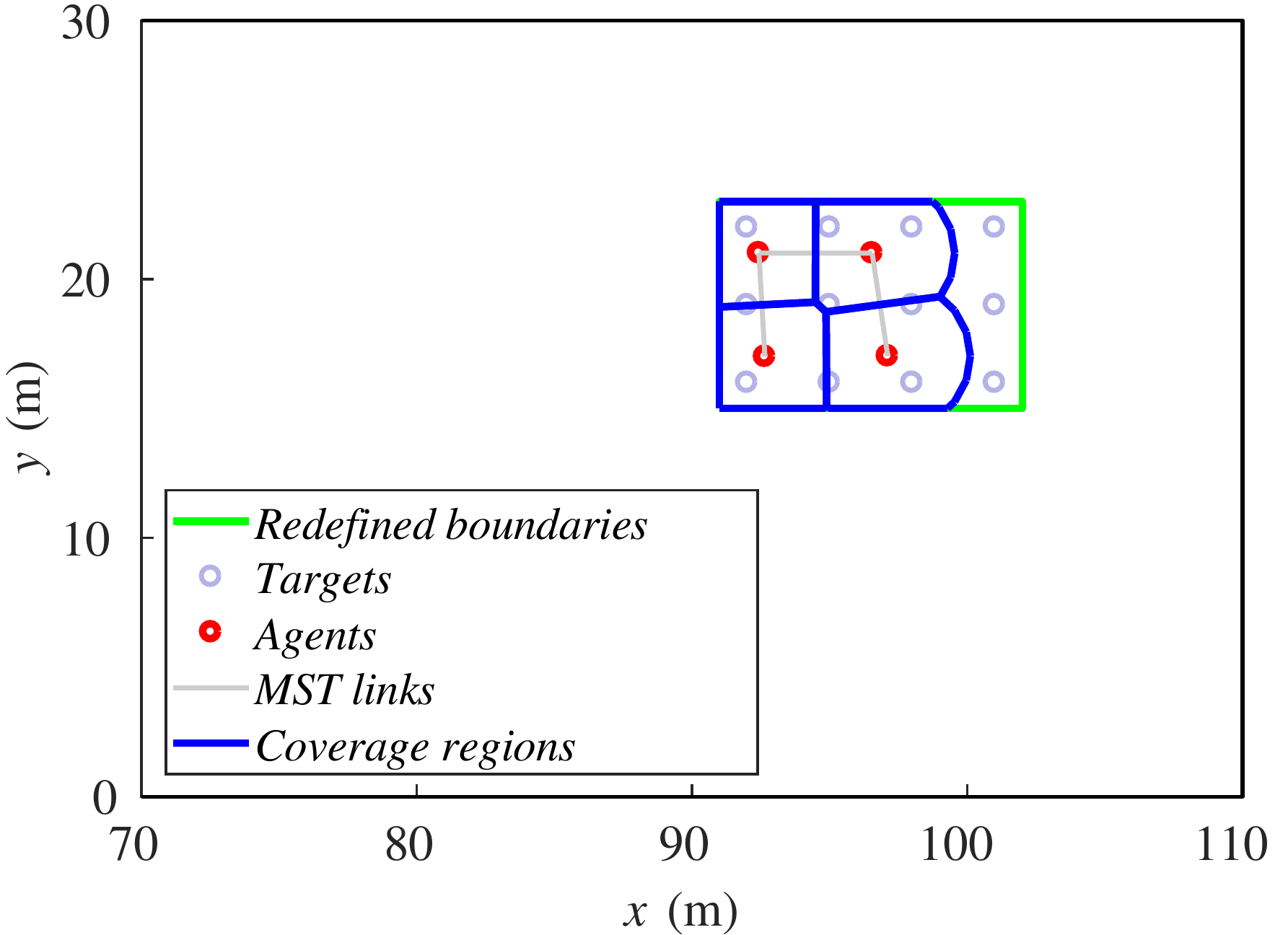}&
		\includegraphics*[width=0.45\textwidth]{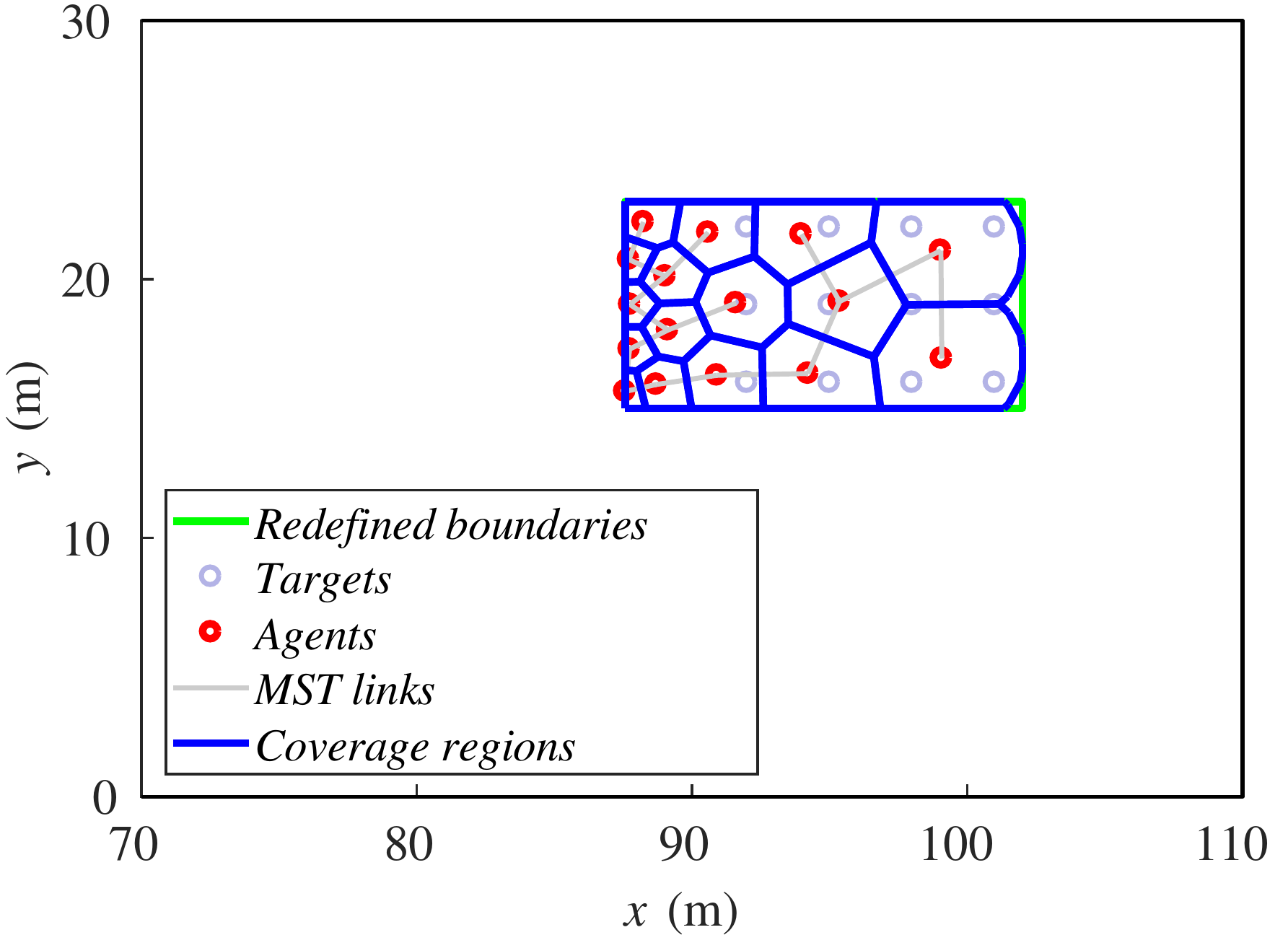}\\
		{\footnotesize \textbf{(a)}} & {\footnotesize \textbf{(b)}}
	\end{tabular}	
	\caption{Final positions of agents and targets for (\textbf{a}) 4 robots and (\textbf{b}) 16 robots.}
	\label{final_pos_rob}
\end{figure}

Contrary to what one could have expected, we observe that, with a higher amount of agents, the method does not perform better, even though they have larger capabilities to cover an area. The mean distance to the center of the formation increases as the number of agents is higher. Observing Figure \ref{final_pos_rob}, one can see the reason of this behavior. When there are more ground robots than needed, the extra agents tend to accumulate at the rear formation. In case that this excess is extreme, the agents are very close and interfere with each other, making their available space to move too short. This prevents them to advance with the formation of targets, and their mean distance to the center of the formation increases.

The agents that lead the motion of the team are those that have direct contact with a boundary perpendicular to the direction of the target formation, i.e., the ones that are at the front and rear. They have a direct modification of their Voronoi and sensing region, implying a quicker reaction than the agents that are in the middle part of the formation. So the accumulation of unnecessary ground robots increases the number of agents in the center of the formation, obstructing the motion of the whole team. In case that many ground robots were needed, the exploration strategy of the aerial team should consider a limit speed appropriate for the features of the system.

\subsection{Tracking Method}
Finally we compare the behavior of the algorithm on the base experiment varying the particular method for the tracking task between the proposed in this paper. The initial positions of the agents are again the positions shown in Figure \ref{init_position} in both cases.

\begin{figure}[H]
	\centering
	\begin{tabular}{cc}
		\includegraphics*[width=0.45\textwidth]{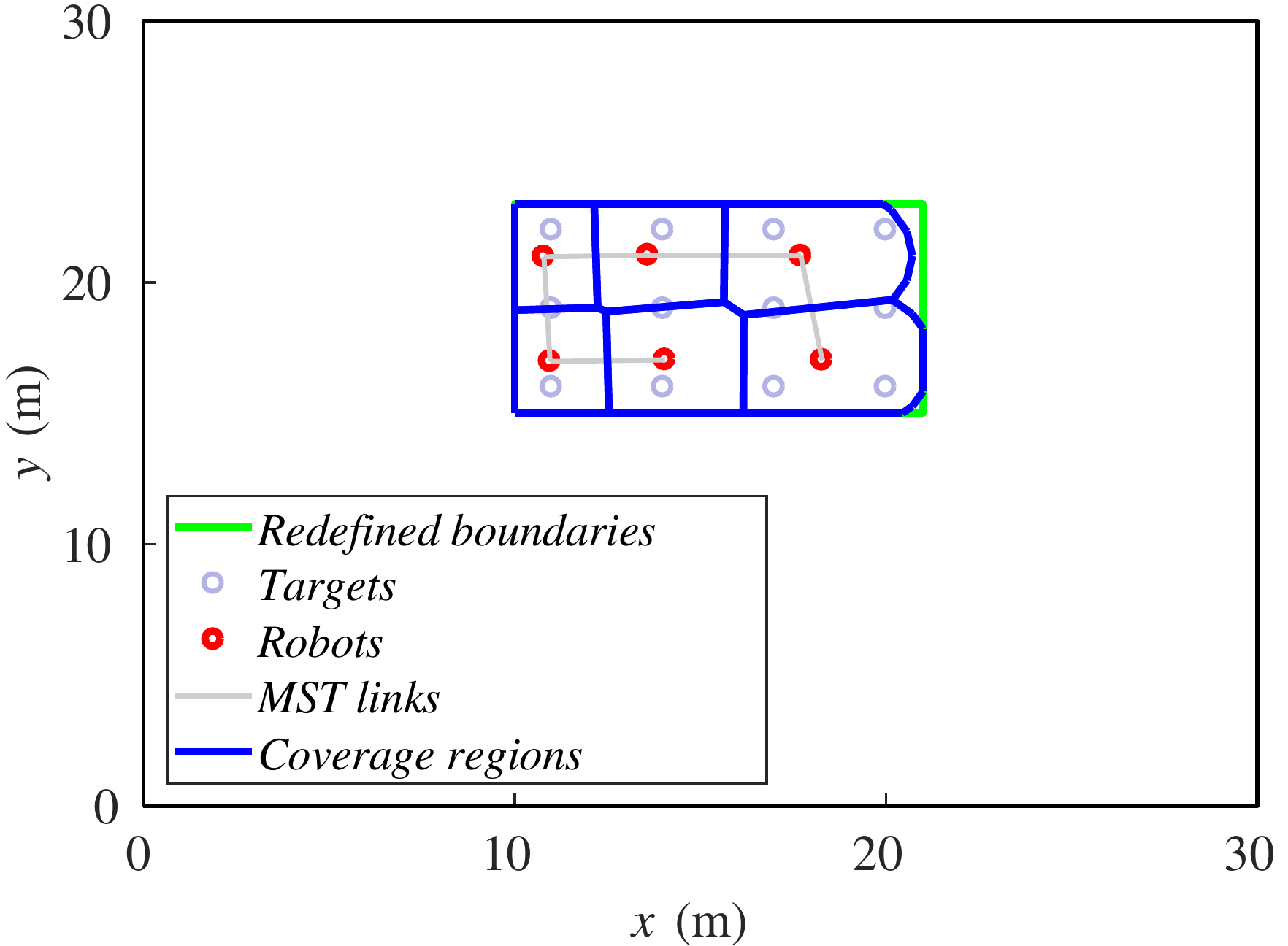}&
		\includegraphics*[width=0.45\textwidth]{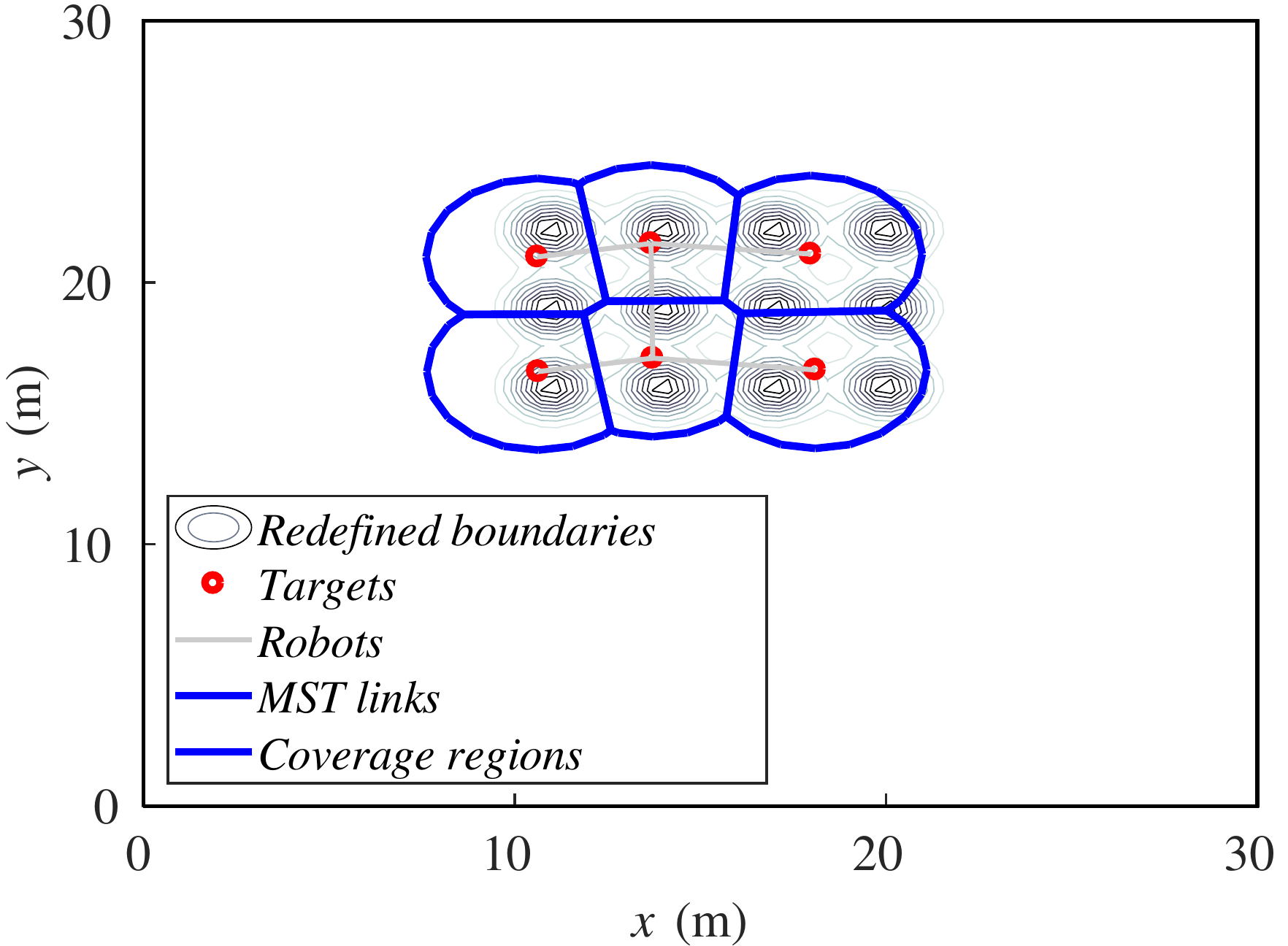}\\
		{\footnotesize \textbf{(a)}} & {\footnotesize \textbf{(b)}}
	\end{tabular}	
	\caption{Final positions of agents and targets for (\textbf{a}) boundaries redefining and (\textbf{b}) importance functions methods.}
	\label{final_pos_met}
\end{figure}

After a 30 step simulation, the final positions of the agents are shown in Figure \ref{final_pos_met}. The distributions of the agents over the formation are very similar. Both alternatives cover the twelve targets when the steady state is reached.

\begin{figure}[H]
	\centering
	\includegraphics[width=0.5\textwidth]{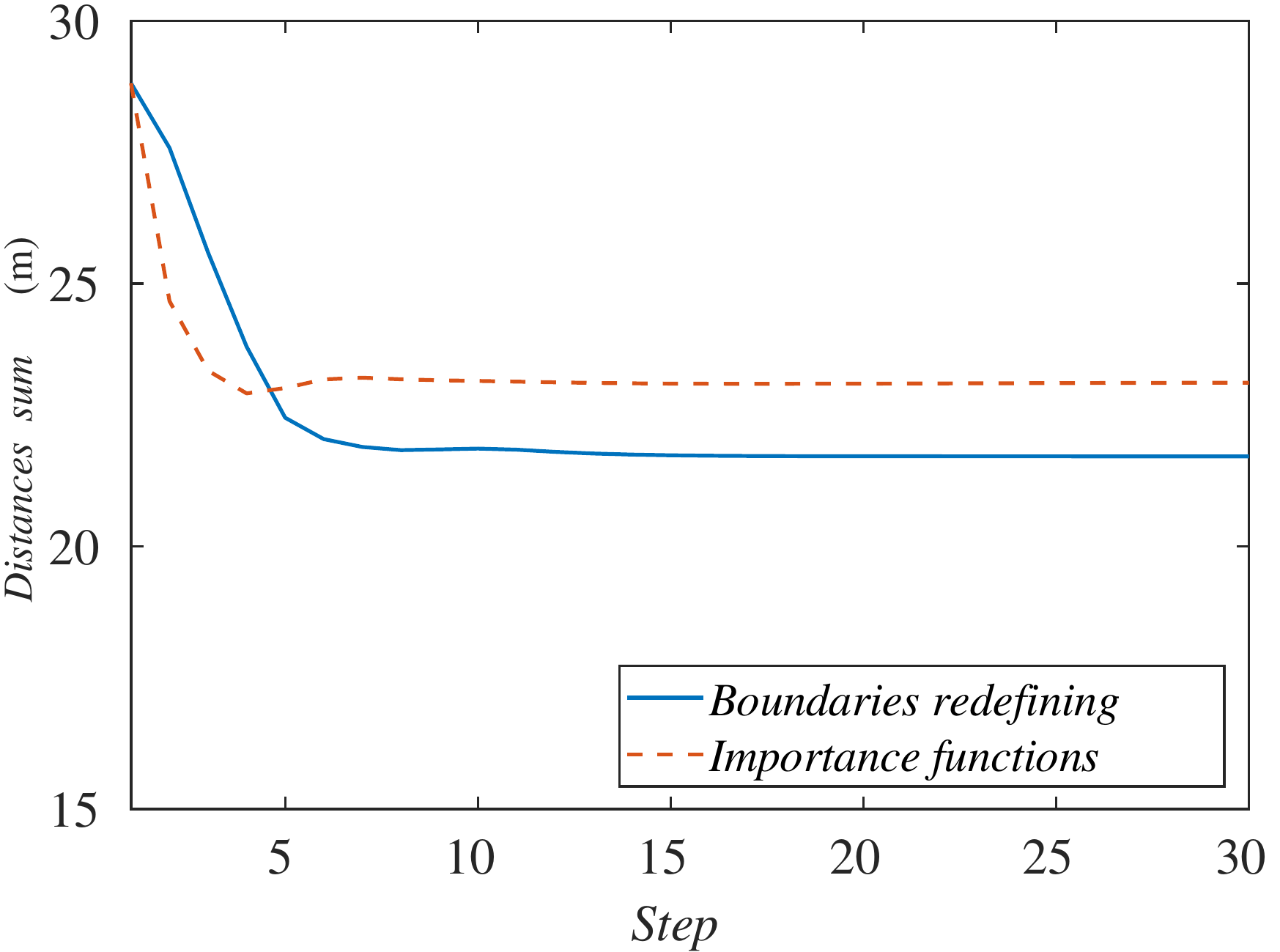}
	\caption{Sum of agents distance to the formation center along the simulation depending on the tracking method.}
	\label{robotdist_met}
\end{figure}

Representing the sum of the agents distances to the center of the formation (Figure \ref{robotdist_met}) we can see that the value of the importance functions method quickly decreases, so its deployment is faster and its response time is lower. On the other hand, the boundaries redefining method achieves a lower value, so the agents in this case are a bit better deployed.

We can conclude then that both methods are suitable for this experiment, but the one redefining the boundaries leads the comparison with a better deployment. In addition, as we saw in Section \ref{tracking_section}, this alternative is more appropriate for targets agglomerations like the ones considered in the scenario discussed in this section.

\section{Realistic Simulations}
\label{realistic_sim_section}

With the aim of getting one step closer to a real implementation of the methods proposed in Section \ref{tracking_section}, we have implemented and tested our methods using a realistic 3D physics simulator specialized in robotics, Gazebo. We have used models associated to real robots --Turtlebots-- which have a differential drive behavior. They are controlled with ROS through specific packages that include the robot description, the robot controller or the navigation stack. Configuring and connecting the set of packages, we are able to send goal positions to the ground agents once we compute the coverage algorithm. Iteratively, Turtlebots execute the Algorithms \ref{importance_algorithm} or \ref{boundaries_algorithm} described in Section \ref{tracking_section}. Between iterations, they navigate to their goal positions using local lower level controllers, taking into account their own motion kinematics restrictions. Since in the higher level methods described in Sections \ref{coverage_section} and \ref{tracking_section} we have not included any collision avoidance method, we incorporate a reactive local method in the Gazebo implementation so that agents do not collide as they navigate.

Lloyd methods, as other distributed strategies, make some assumptions that make the problem more tractable. In particular, in our simulations we have made the following assumptions:
\begin{itemize}
\item Scenarios are obstacle-free. Robots only need to avoid collision with other robots, but not with other objects in the environment.
\item Ground robots can exchange data in one hop with robots within their communication radius.
\item Ground robots share a common orientation, as required by Algorithm 4.2.
\item Targets receive data from the ground robots to know that they have reached their goal destinations. They wait for this confirmation before performing their next motion.
\item Ground robots can also localize themselves, localize their neighbors, and get information of the positions of the targets --here, aerial robots--, by direct measurement, or by multi-hop messages. We use the noise-free \textit{ground-truth} data given by Gazebo. 
\end{itemize}

\begin{figure}[H]
	\centering
	\setlength{\fboxsep}{0pt}%
	\setlength{\fboxrule}{1pt}%
	\fbox{\includegraphics[width=0.65\textwidth]{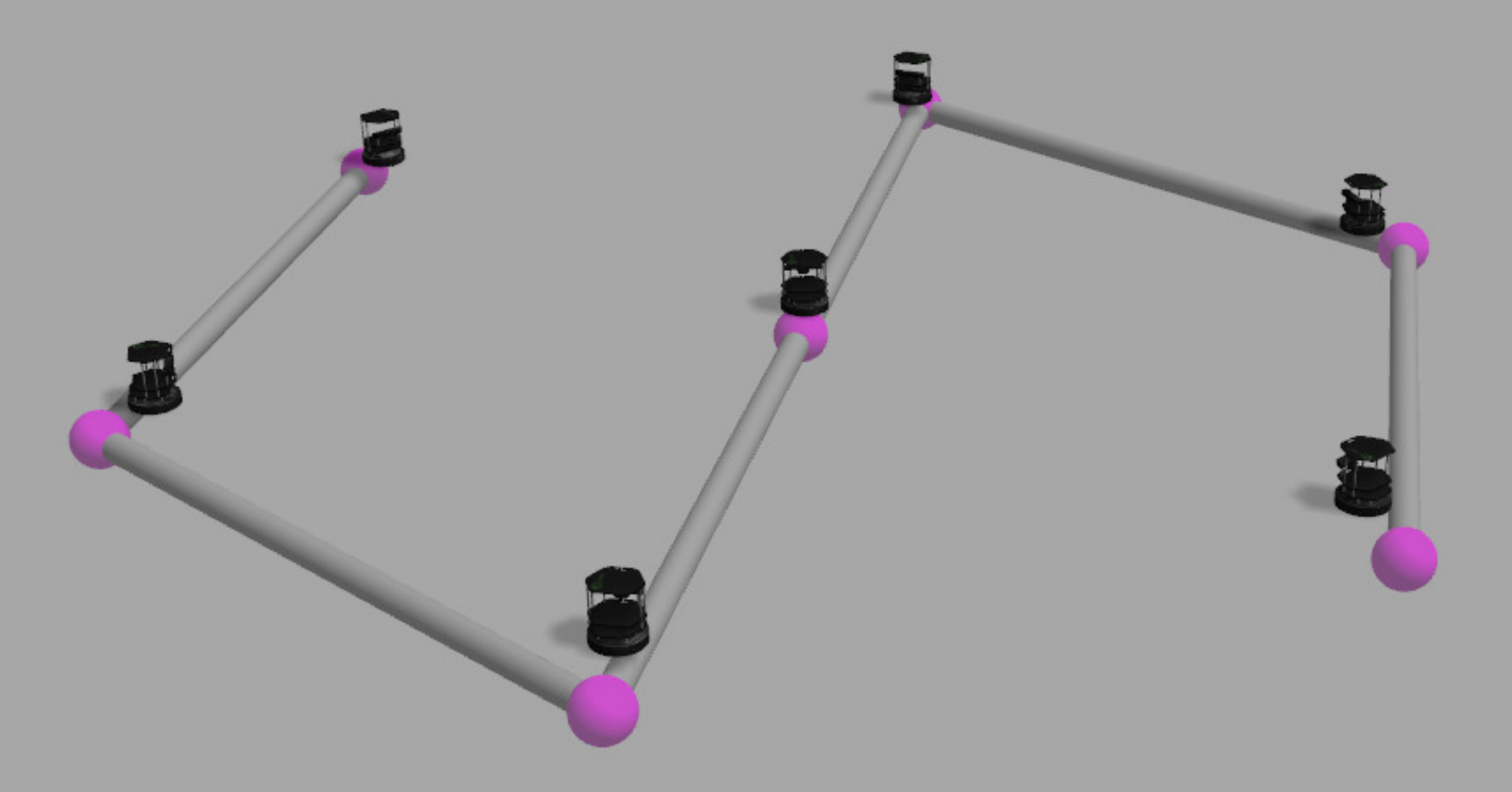}}
	\caption{Final layout in a expansion without targets.}
	\label{Gazebo_expansion}
\end{figure}

We have designed four obstacle-free scenarios in order to examine the behavior of the algorithm in some common situations and some extreme cases where it might fail. The reader can find in \cite{public-videos} four videos with names ``\textit{case\_i}'', where $i$ is a number from 1-4. The videos represent the simulations carried out in this section, in order of appearance. They contain the simulation in Gazebo and the graphs obtained from MATLAB. The first one is a simple expansion without targets or zones more important than others. This is just a coverage task and we do not need to implement any of the tracking alternatives proposed. Figure \ref{Gazebo_expansion} shows the final layout of the ground agents. We have developed a visualization system that helps to understand the movement of each agent in the Gazebo simulation. The black ground robots are the Turtlebots, which are the ones actually running the algorithms. We add some elements like the targets, implemented as aerial robots that are moved discretely. The colored spheres represent the positions of the ground robots and their goal positions in the current simulation step. And the gray cylinders represent the communication tree reached at present step.

\begin{figure}[H]
	\centering
	\setlength{\fboxsep}{0pt}%
	\setlength{\fboxrule}{1pt}%
	\fbox{\includegraphics[width=0.7\textwidth]{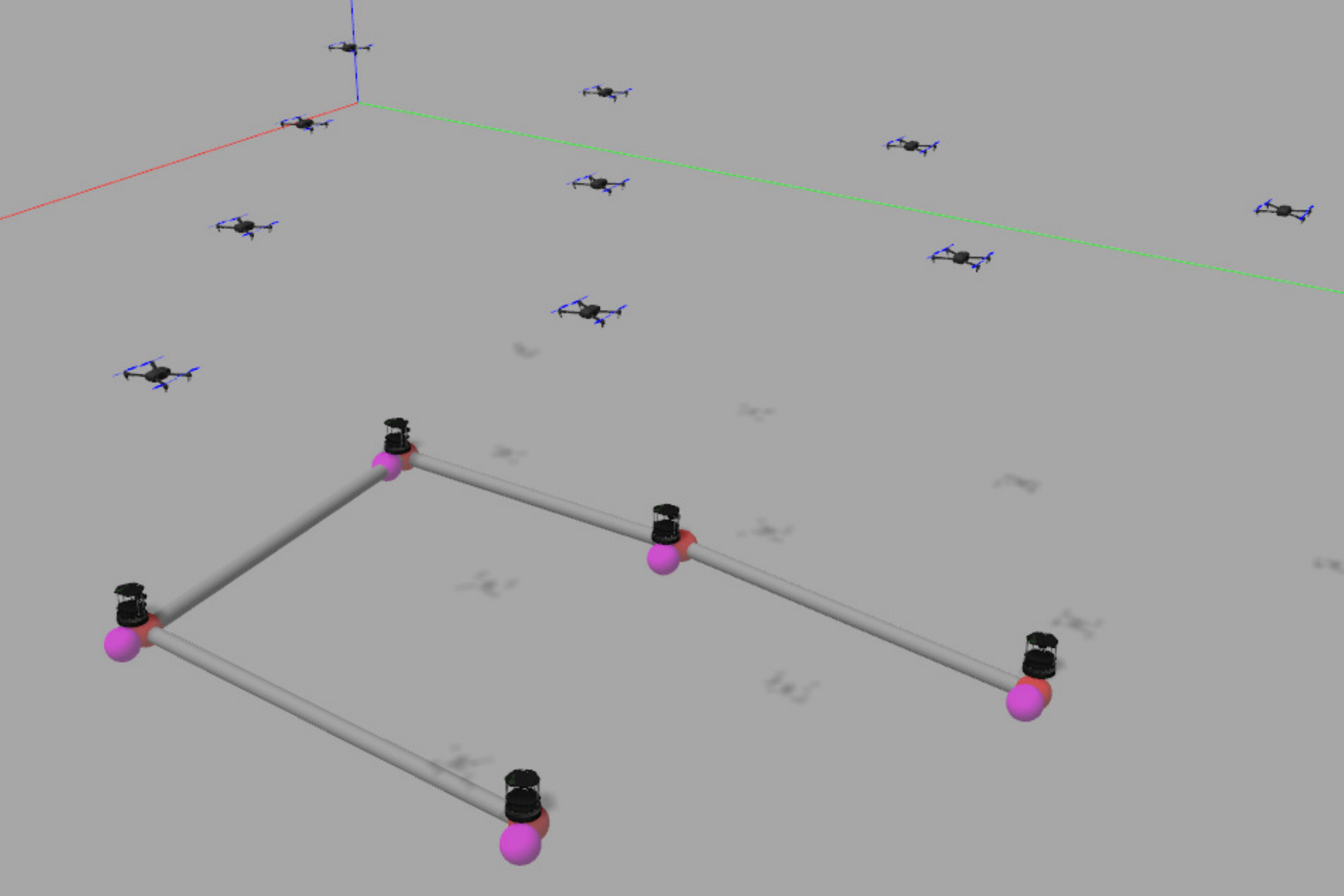}}
	\caption{Final layout in a formation tracking experiment.}
	\label{Gazebo_formation}
\end{figure}

The second experiment consists on five agents implementing the boundaries redefining method covering and tracking a formation of ten quadrotors that fly in a triangular formation. The final disposition of both ground and aerial robots is shown in Figure \ref{Gazebo_formation}.

\begin{figure}[H]
	\centering
	\setlength{\fboxsep}{0pt}%
	\setlength{\fboxrule}{1pt}%
	\fbox{\includegraphics[width=0.7\textwidth]{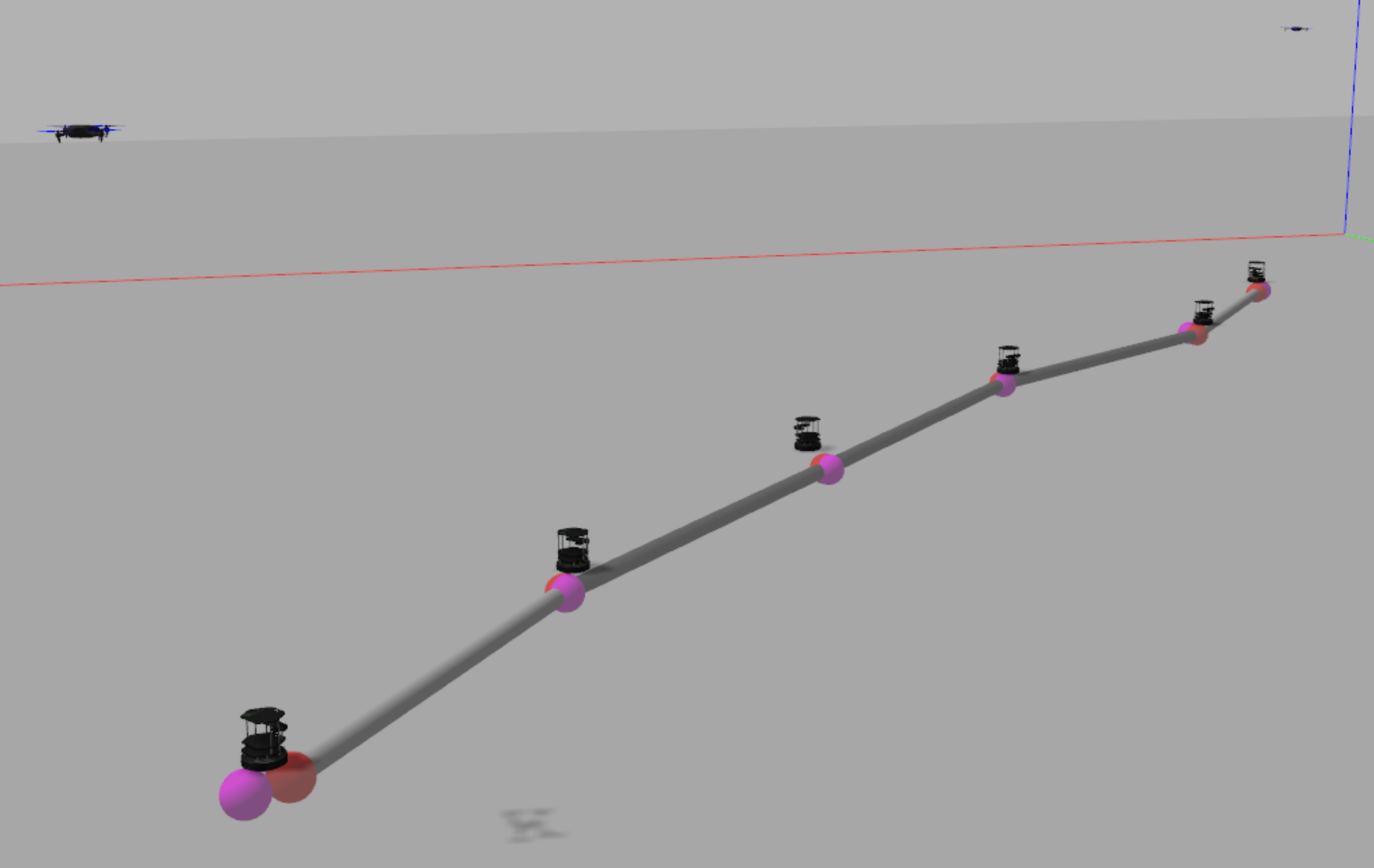}}
	\caption{Final layout in a experiment with two targets separating in opposite directions.}
	\label{Gazebo_separation}
\end{figure}

Third and fourth experiments are implementing the method with the importance functions to track the targets. In the third case the targets start together and separate in opposite directions. The ground team spread in a straight line until the point the communication tree allows them. Figure \ref{Gazebo_separation} shows this final straight formation of the agents.

\begin{figure}[H]
	\centering
	\setlength{\fboxsep}{0pt}%
	\setlength{\fboxrule}{1pt}%
	\fbox{\includegraphics[width=0.7\textwidth]{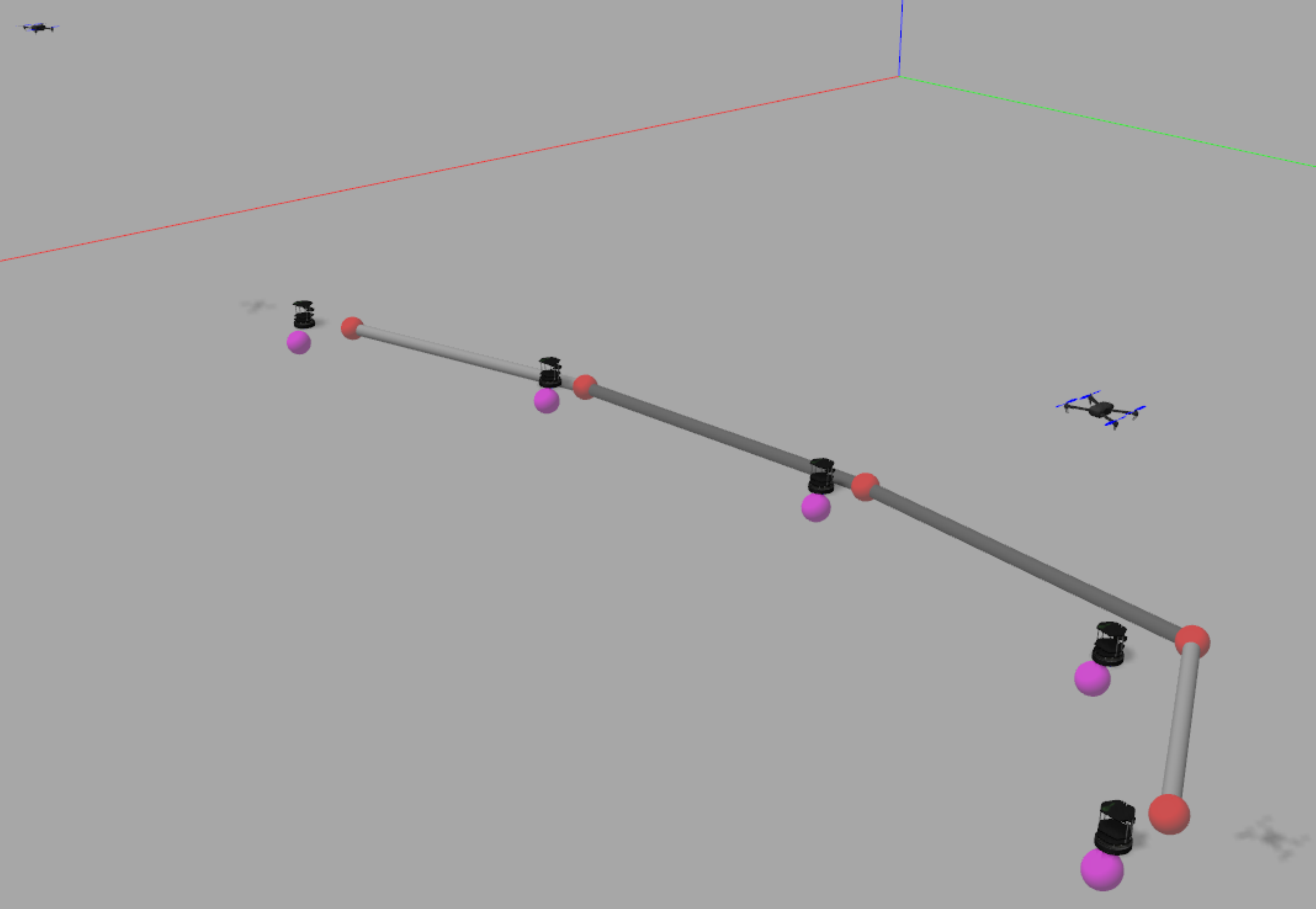}}
	\caption{Final layout in a experiment with two targets following crossed trajectories.}
	\label{Gazebo_cross}
\end{figure}

Finally, in the fourth case two targets describe curve trajectories that cross in the center of the working area. In the first part of the experiment, the ground team covers the aerial targets and track them. During the crossing, the ground agents get too close. Their collision avoidance method works well and prevents them from crashing. However, their redeployment can not be correctly performed because of the agglomeration produced, finally leaving one of the targets uncovered. Figure \ref{Gazebo_cross} shows this fact at the end of the simulation.

\section{Experimental Results}
\label{experiment_section}

In addition to the simulations executed in Sections \ref{simulations_section} and \ref{realistic_sim_section}, we present a real robot experiment in order to analyze the algorithm performance in a more realistic environment. We test the boundaries redefining method as it is the major contribution of this article.

\begin{figure}[H]
	\centering
	\setlength{\fboxsep}{0pt}%
	\setlength{\fboxrule}{1pt}%
	\fbox{\includegraphics[width=0.75\textwidth]{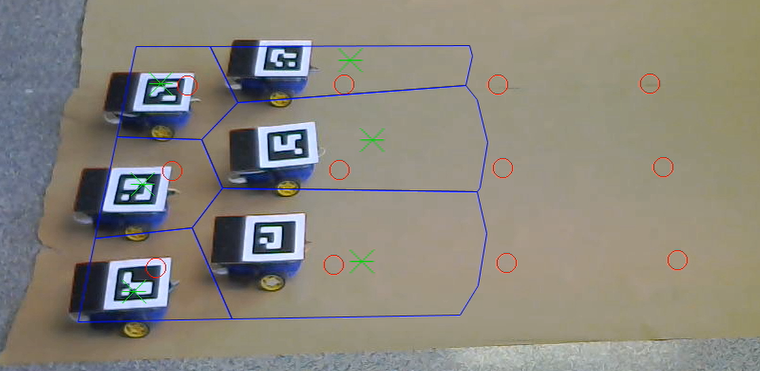}}
	\caption{Initial configuration of the experiment setup.}
	\label{initial_config}
\end{figure}

The experimental tests were carried out with the setup shown in Figure \ref{initial_config}. Six differential--drive robots act as agents, each of them includes a Raspberry Pi 2 Model B, and a driver controller. The localization of each robot is achieved using aRuCo \cite{Aruco2014} markers together with a webcam Logitech C310. So the source of the localization is noisy instead of the ideally precise $ground-truth$ data used in the Gazebo simulation in Section \ref{realistic_sim_section}. Moreover, motions of the robot wheels are affected by real actuation restrictions and by occasional wheel slips. The robots are controlled by a goal-to-goal controller programmed in MATLAB, and a wheel velocity controller running inside each Raspberry Pi with a Python code reading the encoders placed on each wheel and controlling the speed. Commands are sent to the robots by a TCP/IP server opened from MATLAB to each Raspberry Pi, and the Python code sends the voltage messages to the driver controller which powers the motors.

\begin{figure}[H]
	\centering
	\setlength{\fboxsep}{0pt}%
	\setlength{\fboxrule}{1pt}%
	\fbox{\includegraphics[width=0.75\textwidth]{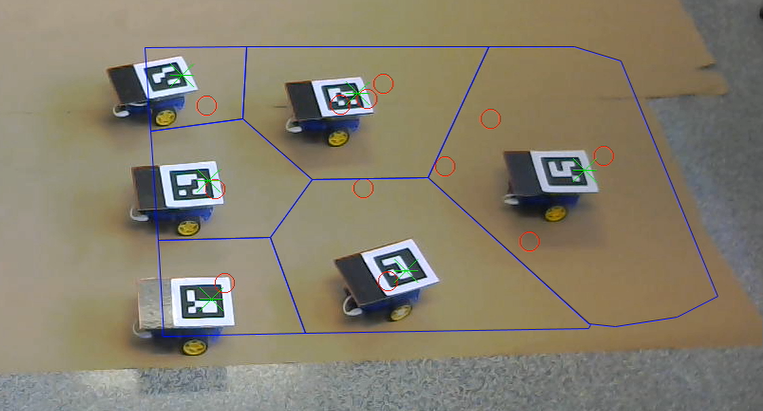}}
	\caption{Setup configuration in a certain step of the experiment}
	\label{experiment_step}
\end{figure}

In Figure \ref{experiment_step} a frame of the experiment conducted can be observed. The video recording this experiment can be found at \cite{public-videos} with name ``\textit{experiment}''. The robots perform the role of agents and the moving targets which should be covered are plotted with red circles. The movement of the robot reach the green asterisks representing the centroids of each coverage region, which are drawn in blue lines. The sensing radius of the agents is set to $s=500$ mm. All of these elements are plotted from the algorithm by homography at each frame of the video.

The movement designed for the targets is composed of two straight sections joined by a semicircle, scheduled beforehand in a 3 by 4 formation. The formation speed is $v=35$ mm/step, much lower than in the simulations. The movement set to the external targets during the curve is scheduled in order to maintain their linear velocity during the curve trajectory. These external agents perform a 750 mm radius curve, while the radius of the internal is 150 mm. The collision between the agents is avoided by narrowing the Voronoi regions by a safety radius of $r_{saf}=75$ mm. 

The control law scheme used in the experiment is the one described in Section~\ref{section_connCtrl}: agents receive goal positions, generated according to Eq.~\eqref{eq_controlLaw1} --green asterisks--. Note the goal positions at each iteration are  the result of computing the centroids of the Voronoi-regions, and adjusting them so as to keep the communication links, as described in Sections~\ref{coverage_section}-\ref{tracking_section}. Note also that the low-cost robots in the experiment  have differential--drive kinematics. Thus, agents run their low level controllers to get close the high level goal positions. 
When the agents are near enough to their goal (we use a distance tolerance $d=65$ mm in our experiments), they stop moving and wait for the remaining agents to finish navigating to their their goals. After that, all the agents start a new iteration of Algorithm~\ref{boundaries_algorithm}, which includes the re-computation of Voronoi regions. 

\begin{figure}[H]
	\centering
	\includegraphics[width=0.65\textwidth]{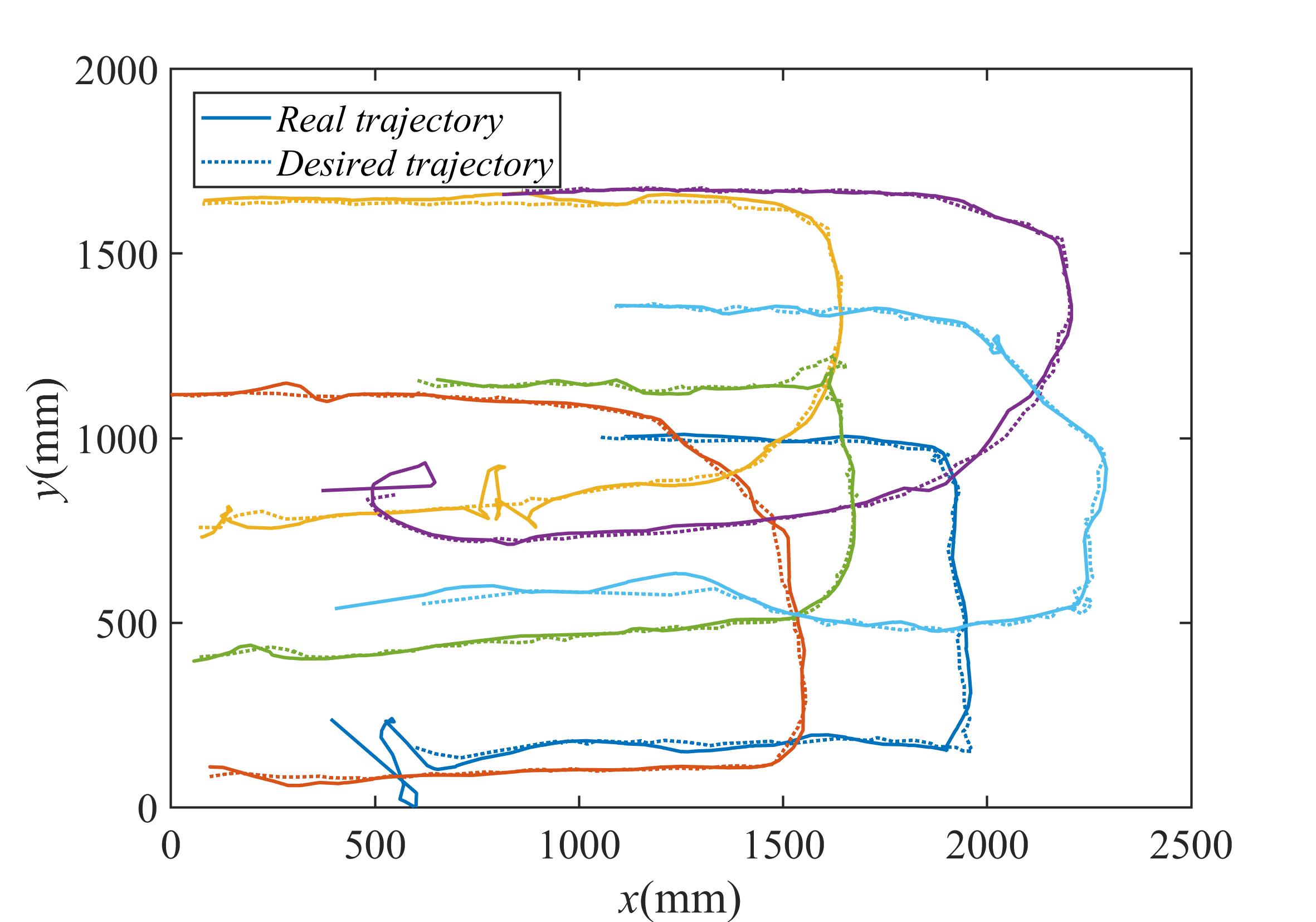}
	\caption{X-Y representation of the consecutive goal positions and the real trajectories of the agents.}
	\label{trajectory}
\end{figure}

Figure \ref{trajectory} represents the real trajectories followed by the different agents --continuous line-- against the set of consecutive goal positions (Eq.~\eqref{eq_controlLaw1}) derived from the algorithm --dotted line--.

\begin{figure}[H]
	\centering
	\includegraphics[width=0.65\textwidth]{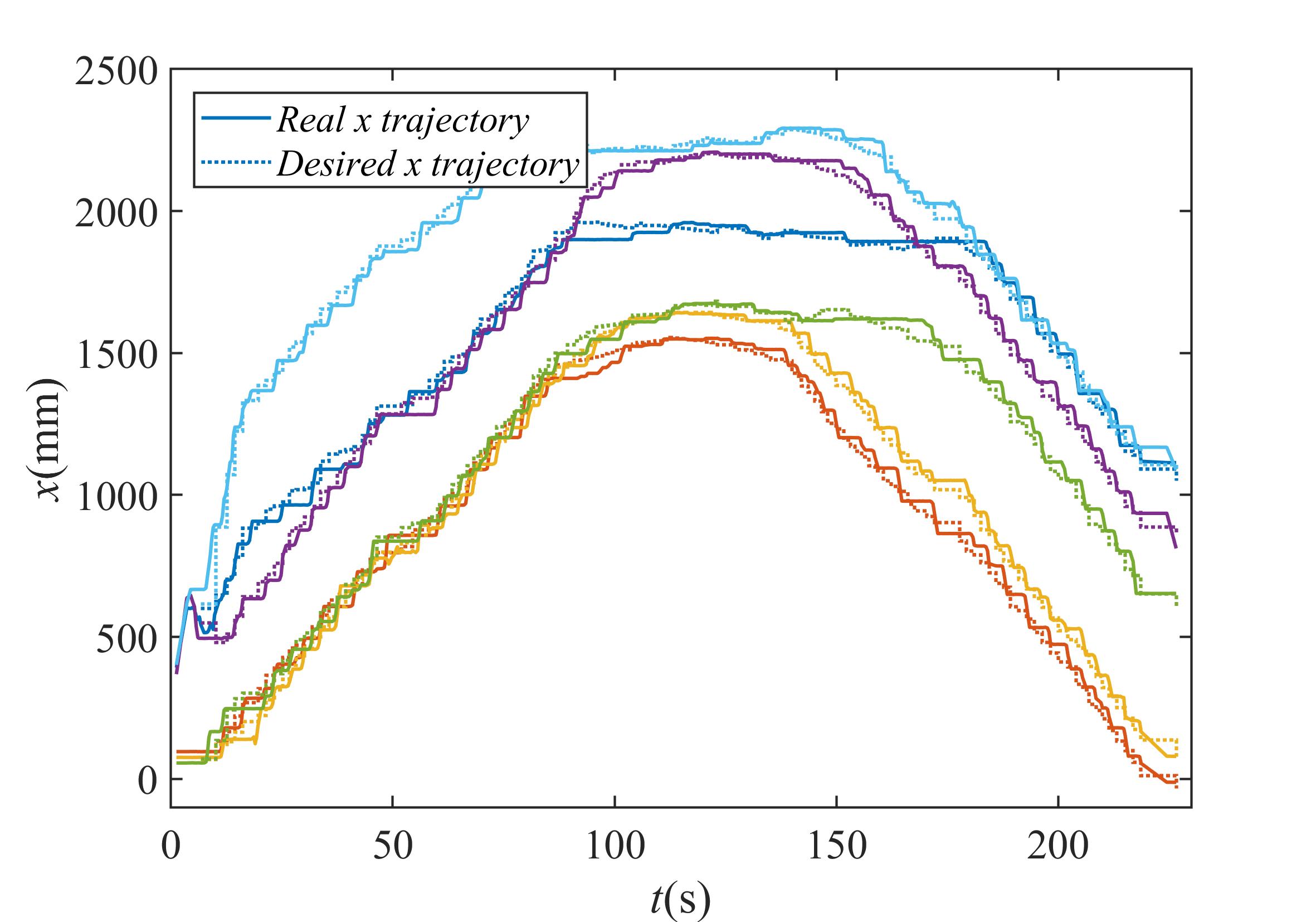}
	\caption{T-X representation of the consecutive goal positions and the real trajectories of the agents.}
	\label{trajectory_t}
\end{figure}

In Figure \ref{trajectory_t} we represent evolution of the X coordinate of the trajectories and the goal positions along the time of the experiment.

Observe how the desired high--level goal positions (Eq.~\eqref{eq_controlLaw1}) remain unchanged during some time. There, the low--level controllers of the agents drive them from their current positions to these new goals. Thus, as it can be seen, the algorithms \ref{importance_algorithm}, \ref{boundaries_algorithm} proposed in this paper can be run by  agents with different kinematic models, as long as they are equipped with local controllers that drive them to the high level goal positions generated by equation~\eqref{eq_controlLaw1}.

\section{Conclusions}
\label{conclusions_section}

We propose and test two strategies to solve the problem of simultaneous coverage and tracking, taking into account sensing and communication limitations, while ensuring that the network remains connected. Both strategies are based on Lloyd methods for static deployment. The first consists of modifying the positions of density functions according to the movement of the targets. This solution has been commonly used in the literature for adapting the static algorithm to tracking tasks. The second strategy iteratively modifies the boundaries of the working area in order to follow the targets while covering their surroundings. As far as we know, this is a novel solution for the problem addressed in this paper.

The density functions solution results to react faster to the targets motion. However, depending on the initial disposition of the agents, some targets may end up uncovered, while the agents concentrate around other targets. In addition, all of the agents need to know the current positions of the targets in order to compute the algorithm.

On the other hand, the boundaries redefining solution performs a slower reaction to the targets motion, but achieves an even deployment over the area of interest. Furthermore, the information about the position of the targets is traduced into the position of the boundaries, and this information is only needed by the agents that have contact with them. So, the communication load is lighter when implementing many agents. A parametric study over the second method remarks the existence of a maximum velocity of the targets that it can follow. We have shown that this limit speed depends on the number of agents and their sensing radius, and does not depend on the communication radius. We have implemented and validated the system on a realistic simulator. We have carried out real experiments with a team of six low-cost robots. We have shown that the methods can cope with measurement noises and that can be combined with several kinematic robot models.
\\

\vspace{6pt} 

\acknowledgments{Supported by Spanish Ministerio de Economía y Competitividad DPI2015-69376-R and Vicerrectorado de Política Científica, Universidad de Zaragoza JIUZ-2017-TEC-01 projects.}

\reftitle{References}

\externalbibliography{yes} 
\bibliography{bibliography}

\end{document}